%% file: main.tex
\documentclass[10pt,twocolumn,letterpaper]{article}

\usepackage[pagenumbers]{iccv} 


\definecolor{iccvblue}{rgb}{0.21,0.49,0.74}
\usepackage[pagebackref,breaklinks,colorlinks,allcolors=iccvblue]{hyperref}
\usepackage{multirow}
\usepackage{bm}
\usepackage{amsfonts}
\usepackage{enumitem}

\def\paperID{*****} 
\def\confName{ICCV}
\def\confYear{2025}

\title{HuGeDiff: 3D Human Generation via Diffusion with Gaussian Splatting}

\author{\parbox{16cm}{\centering
    {\large Maksym Ivashechkin, Oscar Mendez, Richard Bowden}\\
    {\normalsize
    CVSSP, University of Surrey, Guildford, United Kingdom}\\
    {\normalsize \texttt {\{m.ivashechkin, o.mendez, r.bowden\}@surrey.ac.uk}}}%
}

\begin{document}
\maketitle

\input{sections/abstract}
\input{sections/introduction}
\input{sections/related_work}

\input{sections/methodology}
\input{sections/experiments}

\input{sections/limitations}

\input{sections/conclusions}

\input{sections/acknowledgements}

\clearpage

{
    \small
    \bibliographystyle{ieeenat_fullname}
    \bibliography{main}
}

\clearpage \input{sections/suppl} 

\end{document}

%% file: sections/abstract.tex
\begin{abstract}
3D human generation is an important problem with a wide range of applications in computer vision and graphics.
Despite recent progress in generative AI such as diffusion models or rendering methods like Neural Radiance Fields or Gaussian Splatting, controlling the generation of accurate 3D humans from text prompts remains an open challenge.
Current methods struggle with fine detail, accurate rendering of hands and faces, human realism, and controlability over appearance.
The lack of diversity, realism, and annotation in human image data also remains a challenge, hindering the development of a foundational 3D human model.
We present a weakly supervised pipeline that tries to address these challenges.
In the first step, we generate a photorealistic human image dataset with controllable attributes such as appearance, race, gender, etc using a~\emph{state-of-the-art} image diffusion model.
Next, we propose an efficient mapping approach from image features to 3D point clouds using a transformer-based architecture.
Finally, we close the loop by training a point-cloud diffusion model that is conditioned on the same text prompts used to generate the original samples. 
We demonstrate orders-of-magnitude speed-ups in 3D human generation compared to the~\emph{state-of-the-art} approaches, along with significantly improved text-prompt alignment, realism, and rendering quality.
We will make the code and dataset available.
\end{abstract}

%% file: sections/introduction.tex
\section{Introduction}


3D human modeling, including reconstruction, rendering, and generation, plays a more significant role than ever in modern computer vision, supporting applications in augmented/virtual reality, human-computer interaction, gaming and telepresence.
Recent advances in the field, alongside increased computational power, have enabled the creation of photorealistic avatars.
However, the field still faces many challenges, including model generalization, data diversity and privacy (\emph{e.g.}, people appearance), clothing variability, method performance, etc.


Efficient generative 3D human models can greatly benefit the field by enabling appearance transfer, which helps preserve anonymity and increase diversity.
However, such a generative model requires large-scale and diverse datasets, which remain a major challenge.
Most real-world human datasets are constrained by privacy concerns, financial limitations, commercial restrictions, small sample sizes, or demographic biases (limited to the population of the country where the dataset was collected). 
In turn, the lack of data can severely impact the ability of models to generalize.
To mitigate this issue, recent works~\cite{zhuang2024idolinstantphotorealistic3d,yang2025sigmanscaling3dhumangaussian} have begun generating synthetic data or combining existing datasets.
However, the realism of such synthetic data is inherently limited by the quality of the generative models.
While early works~\cite{Rombach_diff,zhang2023adding} often produced artifacts and lacked realism, modern diffusion models~\cite{flux2024,podell2023sdxlimprovinglatentdiffusion} have made significant progress toward generating photorealistic images.
In the literature~\cite{qiu2025LHM}, authors propose collecting samples from public repositories (\emph{e.g.,} YouTube), but the ethical aspects of this are questionable, as no permissions are either sought or granted.

Many existing methods for 3D human reconstruction from a single image rely on SMPL-X UV maps~\cite{zhuang2024idolinstantphotorealistic3d,yang2025sigmanscaling3dhumangaussian,zhang2024e3gen}.
However, this is suboptimal for several reasons.
Firstly, the UV to XYZ mapping is not unique, and secondly, it requires prior access to UV maps, limiting its applicability to real-world settings where such mappings are unavailable.
In contrast, we propose a universal and straightforward approach to map image features directly to 3D structure without relying on predefined UV coordinates.
Compared to related work~\cite{qiu2025LHM,zhang2023globalcorrelated}, our method is more general and easier to apply.

Our method directly tackles key challenges such as data scarcity, generalization, diversity, and reconstruction. We introduce a weakly supervised pipeline that integrates data generation, 3D reconstruction, rendering, and generative modeling. The process starts by producing diverse single-view human images using an off-the-shelf diffusion model. We then reconstruct 3D human representations by lifting image features from arbitrary viewpoints and assigning them to a 3D point cloud using an attention-based mechanism. Finally, we complete the human appearance and close the loop by generating 3D human parameters through a guided diffusion model conditioned on textual prompts.

By leveraging diffusion models for initial data generation, we gain explicit control over diversity in human appearance, including variations in clothing, race, gender, etc., which benefits the training of a generative 3D human model.
In parallel, we introduce efficient architectures for the 3D reconstruction (uplift) stage and point-cloud noise prediction within the diffusion process. Our contributions are summarized as follows:
\begin{enumerate}[noitemsep]
    \item A diverse, high-resolution, photorealistic, AI-generated dataset with corresponding 3D human reconstructions.
     \item An efficient and universal UV-free transformer-based reconstruction method that maps image features to 3D point clouds.
     \item A guided point-cloud diffusion sampler conditioned on text prompts for human rendering via Gaussian Splatting.
\end{enumerate}


%% file: sections/related_work.tex
\section{Related Work}
Most 3D human models are explicitly represented as meshes (\emph{e.g.},~\cite{zhuang2024idolinstantphotorealistic3d,yang2025sigmanscaling3dhumangaussian,zhang2023globalcorrelated,qiu2025LHM,lin2023one,shen2023x,hu2023gaussianavatar,moon2024exavatar} etc.), which provides fast training and inference.
The most widely used mesh representation is SMPL(-X)~\cite{SMPL:2015,SMPL-X:2019}.
Implicit human representations~\cite{xiu2022icon,difu,Pesavento2024ANIMAN}, which use continuous functions (\emph{e.g.}, signed distance fields), are less common due to their high computational cost and slower inference.
However, they are more flexible, template-free, and capable of capturing fine-grained details.

Defining a suitable human model is crucial for 3D reconstruction from images, as it provides strong priors that simplify the task.
Several methods~\cite{yang2025sigmanscaling3dhumangaussian,zhuang2024idolinstantphotorealistic3d,zhang2024e3gen} leverage priors from the SMPL-X framework by mapping UV coordinates to 3D mesh vertices.
A common approach involves extracting image features using a pretrained model (\eg, Sapiens~\cite{khirodkar2024sapiens}), projecting them into UV space, and uplifting the UV features to 3D meshes via the pre-defined mapping.
Other approaches, such as LHM~\cite{qiu2025LHM} and GTA~\cite{zhang2023globalcorrelated}, utilize transformer-based architectures.
For instance, LHM adopts a multimodal transformer~\cite{esser_scaling} tailored for human heads, while GTA leverages a triplane representation.
Point-E~\cite{nichol2022pointegenerating3dpoint} of Nichol~\etal proposes an alternative method which combines CLIP~\cite{Radford2021LearningTV} embeddings for integrating text prompts and image features.

Supervision in reconstruction models often involves rendering the 3D output and computing losses against ground-truth images.
Earlier works~\cite{hmrKanazawa17,simplify} used mesh-based rendering, which is fast but limited by mesh resolution.
NeRF-based methods~\cite{anerf,weng2022humannerf,shao2023tensor4d} model continuous color and density fields, providing strong generalization and high fidelity.
Although they are computationally expensive and slow to train and run, even with faster variants~\cite{hnerf,weng_humannerf_2022_cvpr,yu2022plenoxels,barron2023zipnerf,Chen2022ECCV}.
The recently proposed 3D Gaussian Splatting (3DGS)~\cite{kerbl3Dgaussians} introduces a discrete set of Gaussian primitives, enabling real-time rendering while achieving quality comparable to or better than NeRF-like approaches.

Training data for 3D human models typically include multi-view images to resolve single-view ambiguities.
Some of the largest multi-view human datasets include MVHumanNet~\cite{xiong2024mvhumannet}, THuman~\cite{tao2021function4d,zheng2022structured,deepcloth_su2022}, Human3.6M~\cite{human3.6m}, and ActorsHQ~\cite{isik2023humanrf}.
However, limited diversity and control in human appearance have led to a growing interest in generating synthetic datasets or augmenting real ones.
For example, IDOL~\cite{zhuang2024idolinstantphotorealistic3d} of Zhuang~{\etal} proposes human image generation, and Champ~\cite{zhu2024champ} augments the data to simulate multiple views using generative models conditioned on single-view images and meshes.
SIGMAN~\cite{yang2025sigmanscaling3dhumangaussian} and LHM~\cite{qiu2025LHM} combine various public real-world and synthetic datasets to improve generalization.

An obvious solution is to approach 3D human generation from a text-prompt as a two-stage process~\cite{liu2023humangaussian,liao2024tada,cao2024dreamavatar,huang2024dreamwaltz-g,yi2023gaussiandreamer,cao2024dreamavatar,DreamHuman,Kolotouros_instant}, where multi-view images are generated first by Stable Diffusion~\cite{Rombach_diff} or ControlNet~\cite{zhang2023adding}, and then a 3D model is fitted to them.
However, since many approaches build on older diffusion models, these methods lack realism and can take too long to generate.
Methods such as~\cite{gsm,zhang2024e3gen,hong2023evad,yang2025sigmanscaling3dhumangaussian} generate 3D humans in a zero-shot manner, but they lack appearance control. 
The SMPLitex~\cite{casas2023smplitex} has both appearance control and instant generation from text, but generates only UV texture maps.
Gong~{\etal} in~\cite{gong2024text2avatar} and Fu~{\etal} in ~\cite{Fu2023Textguided3H} struggle to render detailed hands and faces.
Human3Diffusion~\cite{xue2024human3diffusion} and PSHuman~\cite{li2024pshuman} provide only an image-conditioned 3D human generation framework.



%% file: sections/methodology.tex
\section{Methodology}
\begin{figure*}
    \centering
    \includegraphics[width=0.99\linewidth]{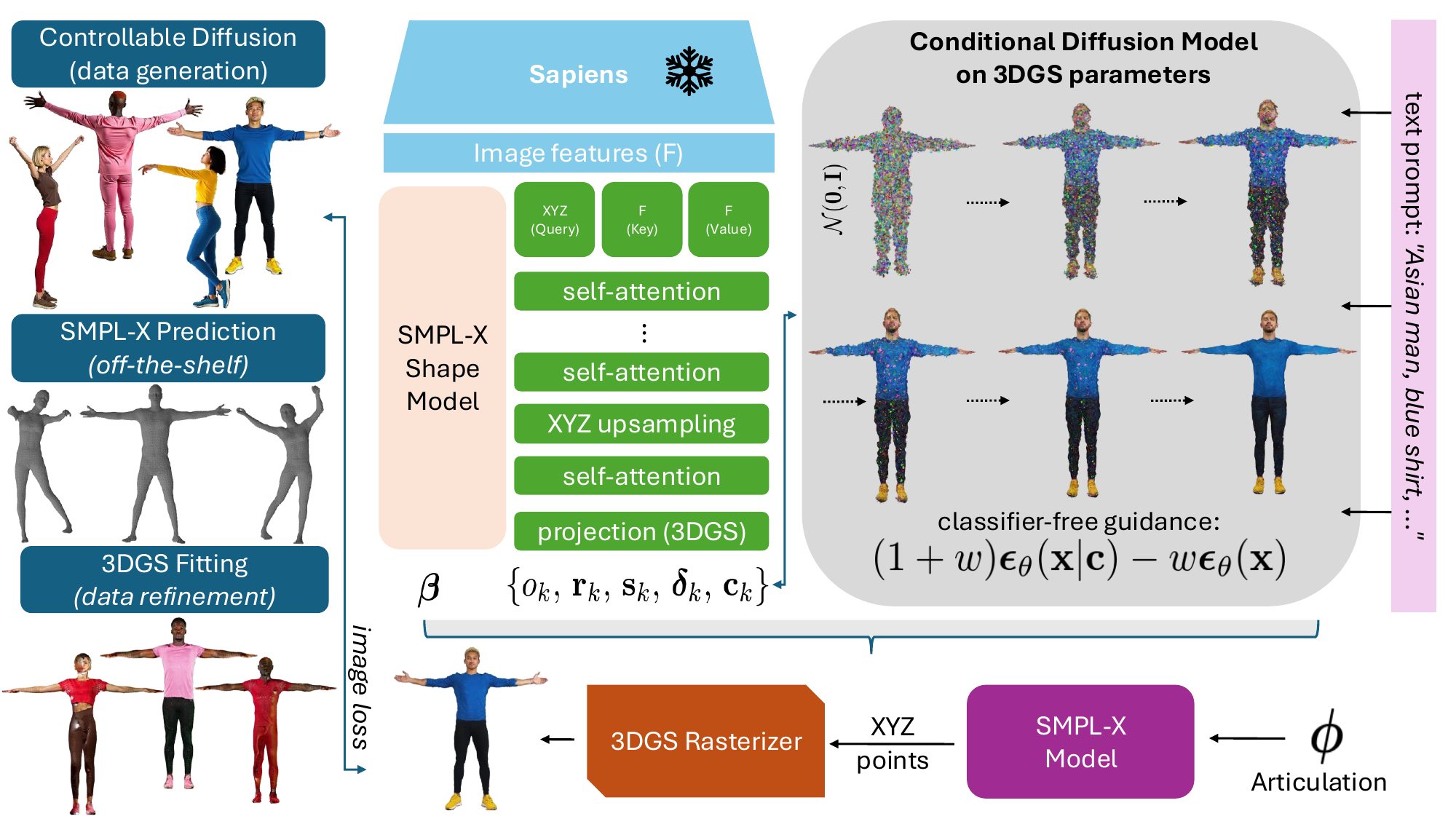}
    \caption{{\bf Overview of the proposed pipeline}. The left side shows the data generation part, the middle part is the 3DGS reconstruction from image features. The right side highlights the text-conditioned diffusion model. }
    \label{fig:pipeline}
\end{figure*}
Our pipeline is illustrated in Fig.~\ref{fig:pipeline} and consists of several key stages.
First, we generate a dataset of diverse humans and perform an initial reconstruction by fitting the SMPL-X parameters to each image.
Next, we train an efficient transformer-based reconstruction model that estimates the 3DGS parameters, anchored to the canonical SMPL-X mesh, using Sapiens features extracted from the image.
Finally, we train a text-conditioned guided diffusion process on 3DGS parameters that learns the human appearance distribution.
We describe each component in detail below.


\subsection{Dataset generation}
\label{sec:dataset_gen}
We employ the FLUX diffusion model~\cite{flux2024} to generate approximately 10,000 diverse human images, conditioned on text prompts describing attributes such as race, gender, hair color, and clothing.
Compared to alternatives like SDXL~\cite{podell2023sdxlimprovinglatentdiffusion} or ControlNet~\cite{zhang2023adding}, FLUX produced the most photorealistic and prompt-aligned results.
Each generated image is segmented using SAM-2~\cite{ravi2024sam2} to remove the background, and then processed with Sapiens~\cite{khirodkar2024sapiens} to extract image features.
We estimate initial SMPL-X parameters using ExPose~\cite{ExPose:2020}, and refine hand poses with HaMeR~\cite{pavlakos2024reconstructing}.
Since ExPose reconstructions are often misaligned, we initialize our Gaussian splatting model with the coarse mesh and jointly optimize SMPL-X and 3DGS parameters by minimizing rendering loss with respect to the original image.
Fitting 3DGS to a single image is highly under-constrained, so we regularize the process by setting all opacities to 1 and aligning rotations perpendicular to surface normals. This enables fast and stable optimization.


\subsection{Single Image Reconstruction}
After obtaining reliable SMPL-X parameters, our next step is to create a generalizable model that reconstructs a 3D human from a single image in terms of 3DGS parameters.
For image $\mathbf{I}$, the human reconstruction is represented by SMPL-X pose vector $\bm\phi$, shape vector $\bm\beta$, and set of Gaussian primitives $\mathcal{G} = \{o_k, \:\mathbf{r}_k, \:\mathbf{s}_k, \:\bm{\delta}_k, \:\mathbf{c}_k \:\:|\:\: k = 1,\dots,N \}$ of $N$ points anchored to SMPL-X model $\mathcal{\theta}$.
The 3DGS parameters ($o, \mathbf{r}, \mathbf{s}, \mathbf{c}, \bm{\delta}, \mathbf{c}$) are attached to canonical point set  $\mathbf{X}_0 = \mathcal{\theta}(\bm\phi_0, \bm\beta_0) \in \mathbb{R}^{N\times3}$, and correspond to opacities, rotations, anisotropic scales, colors, and offsets to the canonical pose coordinates ($\mathbf{X}_0$), respectively. 

To reduce the complexity of estimating this high-dimensional parameter space from a single image, we make the following assumptions.
First, the exact human articulation ($\bm\phi$) is less critical for our task than accurate appearance modeling.
\emph{Off-the-shelf} estimators, such as ExPose or OSX~\cite{lin2023one} and HaMeR for hands, can already estimate an approximate human articulation from a single image.
Second, since all points lie on the mesh surface, we assume, without loss of generality, that they are non-transparent (${o_k = 1}$), as transparency would introduce holes in the mesh.
Furthermore, placing rotations perpendicular to the mesh normals implies better alignment with surface geometry, spatial consistency, and higher robustness to various articulations. Finally, the formalized problem is the following:
\begin{equation}
\mathcal{H} (\mathbf{I}) \sim (\mathcal{G}, \bm\phi, \bm\beta) \sim (\{\:\mathbf{s}_k, \:\bm{\delta}_k, \:\mathbf{c}_k\}, \bm\beta),
\end{equation}

We leverage a pre-trained foundation Sapiens model for image processing that has a large human prior and good generalization.
The Sapiens model returns image features $\mathbf{F}$ as the spatial grid $H^\prime \times W^\prime$ with $C$ channels from the input image $\mathbf{I}$ of $H \times W \times 3$.
The next step is to map the $M = (H^\prime \cdot W^\prime)$ feature cells to the $N$ mesh points, which is a non-trivial problem that requires assigning 2D image-based information to a 3D space.
Previous work has approached this problem using the UV template SMPL-X mapping~\cite{yang2025sigmanscaling3dhumangaussian,zhuang2024idolinstantphotorealistic3d}, which determines the correspondence of the 2D UV coordinates to the 3D model using standard graphics techniques. 
However, taking inspiration from template-free methods~\cite{zhang2023globalcorrelated,qiu2025LHM}, we train an efficient transformer-based model to map image features to the point cloud.
To preserve high-frequency details and maintain efficiency, we design a lightweight cross-attention module between canonical points and image features. The query is a canonical set of points ($\mathbf{X}_0$) and keys with values are image features ($\mathbf{F})$. 
We add a linear projection $l$ along with positional encoding to match the same dimensionality ($d$) of key ($\mathbf{K}$) and query ($\mathbf{Q}$) and reshape the tensors to allow multi-head attention ($h$ -- number of heads). The value ($\mathbf{V}$) of cross-attention is image features split among $h$ heads:
\begin{equation}
 \mathbf{Q} = l_{\text{PE}}(\mathbf{X}_0) \in \mathbb{R}^{N \times h \times 1 \times d},
\end{equation}
\begin{equation}
\mathbf{K} = l_\text{PE} (\mathbf{F}) \in \mathbb{R}^{1\times h \times M \times d},
\end{equation}
\begin{equation}
\mathbf{V} = \text{\emph{split}}(\mathbf{F}, h) \in \mathbb{R}^{1\times h \times M \times \frac{C}{h}}.
\end{equation}

The attention weights are computed by applying a softmax to the multiplication of the query and the key matrices.
Afterwards, the per-point features ($\mathbf{F}_X$) are computed as follows:
\begin{equation}
    \mathbf{F}_X = \text{softmax}(\mathbf{Q} \, \mathbf{K}^\top) \, \mathbf{V},
\end{equation}
In our derivation, we assume broadcasting of tensors for dimension value ``1'' (similarly to PyTorch~\cite{pytorch}).
Additionally, the transpose operator ($^\top$) is applied only to the last two dimensions.
The full description of this equation, along with subsequent ones, can be found in the Supplementary Materials.



The spatial dimensions of the feature grid $(H^\prime, W^\prime)$ (last output of the  ViT~\cite{dosovitskiy2020vit} encoder) normally do not exceed $64\times 64$, and the number of feature channels are 2048 in the largest Sapiens model.
The main computational bottleneck arises from the number of points $N$, which significantly increases the processing cost.
As a result, the final complexity is $\mathcal{O} \big(N\,h\,M\,\max(d,\frac{C}{h})\big)$.
The number of points $N$ is a scalable parameter that influences the quality of the reconstruction, but it must be handled with care to manage the computational demands.
To address this, we introduce an upsampling module that enables the mapping of image features to the anchor set of mesh points, and further enhances the output fidelity with a denser point cloud.

Analogously to the assignment of image features to a point cloud, we design a point upsampling mechanism based on cross-attention.
Let $\mathbf{X}^\prime_0 \in \mathbb{R}^{n\times 3}$ be the subsampled point set (\emph{i.e.}, $n < N$), and $K \in \{1,\dots,n\}^{N\times k}$ be a matrix of $k$ nearest neighbor indices of $\mathbf{X}_0$ to $\mathbf{X}^\prime_0$.
The upsampling transformer takes the full point cloud $\mathbf{X}_0$ as a query, a subsampled point cloud indexed with a neighbors matrix as keys, and associated features as values. The multihead cross-attention is then defined as:
\begin{equation}
    \mathbf{F}_{X_0} \leftarrow \text{softmax}\big(l_\text{PE}(\tilde{\mathbf{X}}_0)l_\text{PE}(\tilde{\mathbf{X}}^\prime_0[K])^\top\big) \: \mathbf{F}_{X^\prime_0}[K],
\end{equation}
where the point set $\tilde{\mathbf{X}}$ indicates expanded $d$-dimensional embeddings of the 3D point cloud $\mathbf{X}$ to the shape ($N \times h \times 1 \times d$).
For $\tilde{\mathbf{X}}^\prime_0$ and $\mathbf{F}_{X^\prime_0} \in \mathbb{R}^{n\times h \times 1 \times f}$, the operator $[\dots]$ means indexing the tensor with indices $K$ of shape $(N \times k)$.
For instance, following the indexing behavior in PyTorch, the output is $\mathbf{F}_{X^\prime_0}[K] \in \mathbb{R}^{N\times h \times k \times f}$.

Connections among neighboring points in graph-structured point clouds are essential for local feature propagation and for adding high-frequency detail.
The graph-convolutional methods like GCN~\cite{Kipf:2016tc} or GAT~\cite{velikovi2017graph} improve the local context within neighborhoods. However, these approaches yielded limited improvement in experiments.
Consequently, we designed a self-attention module for point clouds, where the keys incorporate both feature embeddings and relative positional distances to neighboring points.
Let $K^\prime$ be the neighbors indices of $\mathbf{X}$ to itself ($\mathbf{X}$), then features are updated as follows:
\begin{equation}
    \mathbf{K} = l_\text{PE}(\mathbf{F}_X[K^\prime]) + l_\text{PE}(\tilde{\mathbf{X}}_0 - \tilde{\mathbf{X}}_0[K^\prime]),
\end{equation}
\begin{equation}
    \mathbf{F}_X \leftarrow \text{softmax}(l_\text{PE}(\mathbf{F}_X)\mathbf{K}^\top) \: \mathbf{F}_X[K^\prime].
\end{equation}


{\bf 3DGS Regression}. The 3DGS parameters are estimated in the final layers of our model.
After upsampled point features are refined via the self-attention module, we apply a set of linear layers to regress the colors, scales, and displacements relative to the canonical mesh.
Each output is constrained to a pre-defined range to limit and regularize 3DGS,~\emph{e.g.,} prevent excessively large scales or displacements.
For finer structures, such as the head and hands, we apply stricter constraints due to their smaller spatial extent.
To enforce those constraints, we use a sigmoid for RGB colors and sinusoidal normalization for scales and displacements.
For displacements and scales, we found that sigmoid can suffer from vanishing gradients, leading 

{\bf Multi-view Regularization}.
Estimating a full human appearance from a single-view or partial observation is inherently ambiguous, as many plausible reconstructions exist.
Without sufficient regularization and constraints, neural models tend to eventually overfit to the training data.
In the context of a 3D human appearance from a single image, we observe that after a few training epochs, the model starts mirroring the front-facing appearance to the back.
For example, resulting in artifacts such as ``Janus faces'',~\emph{i.e.}, a face appearing on both the front and back of the head. 
To mitigate this, prior works (\emph{e.g.,}~\cite{zhuang2024idolinstantphotorealistic3d,kirschstein2025avat3r}) leverage~\emph{off-the-shelf} generative models conditioned on the input view to obtain multi-view supervision.
In contrast, we eliminate reliance on the external frameworks by implicitly utilizing a human 3D prior.
Specifically, we train a separate text-conditioned reconstruction model by only replacing the image features $\mathbf{F}$ with text-prompt encodings.
Although this model tends to generate an ``average'; human appearance, it is capable of hallucinating high-detail textures.
We render its output from four canonical views (front, back, left, right) and use these renderings as multi-view supervision for regularizing the feature-based model.

{\bf Shape Model}. The SMPL-X shape coefficients $\bm\beta$ are predicted using an MLP that takes both spatial and averaged features.
In particular, the feature matrix $\mathbf{F}$ is first averaged along height and width dimensions, and then separately averaged along the channel dimension (C).
The resulting features are concatenated and flattened into a single vector. 
This provides the model with both spatially-aware and globally-aggregated information, enabling it to estimate an approximate human body shape.
For text-guided models, the shape is predicted directly from the text encoding.

\subsection{Gaussian Splatting Diffusion}
The final stage of our pipeline generates 3D human parameters from the same text prompts used to guide the image diffusion model.
We adopt diffusion models for this task, as they have demonstrated~\emph{state-of-the-art} performance in generative modeling~\cite{Dhariwal_diffusion,Ho_diffusion}, outperforming earlier approaches such as VAEs~\cite{Kingma2014} and GANs~\cite{goodfellow2014generative}.
 
Most approaches employ latent diffusion models~\cite{latent_diff_model} to accelerate training and inference.
For instance, methods~\cite{NEURIPS2022_40e56dab,DiffGS,chen_learning,lan2024ga} encode point clouds via point-based models such as PointNet(++)~\cite{pointnet,pointnet2}, PointTransformer~\cite{zhao2021point}, DGCNN~\cite{dgcnn}, PointConv~\cite{wu2019pointconv}, etc., and the noise predictor is typically a U-Net~\cite{unet}. 
The trade-off for denoising the latent space is a potential quality reduction in generation, because of error accumulation within the encoder-decoder and diffusion process.
To ensure the best quality generation, we apply our diffusion model to the 3DGS parameters,~\emph{i.e.}, scales, displacements, and colors, estimated from the previous steps of the pipeline.
The U-Net 1D works on sequence data, which requires reordering of the point cloud.
With a human point cloud derived from the SMPL-X mesh, there is no reordering to sort points consecutively.
Therefore, we make the following changes. Firstly, we replace ``Conv1D'' with a lightweight MLP that learns weights to combine the $k-$ nearest neighbor features of a point.
Secondly, we replace ``Conv1D'' downsampling and ``Upsample'' layers with multi-head attention on the nearest neighbors, the same  as in the reconstruction model.


To stabilize training and improve the denoising process, we normalize our parameters to the $[-1, 1]$ range.
To rescale the 3DGS parameters, we use this formula $(2\frac{x - x_{\min}}{x_{\max} - x_{\min}} - 1)$, and the reverse $(\frac{x+1}{2}(x_{\max} - x_{\min}) + x_{\min})$ to scale to the original range.
We exploit classifier-free guidance~\cite{Ho2022ClassifierFreeDG} to control the trade-off between adherence to the input text prompt and generative diversity.
By adjusting the guidance scale, we can bias the model toward stronger semantic alignment or allow for more variation in the output.







%% file: sections/experiments.tex
\section{Experiments}

\input{tables/generation_speed_comparison}
\input{tables/score_image_evaluation}


\subsection{Implementation Details}
The original SMPL-X mesh contains around 10,475 mesh vertices. However, such a low resolution leads to poor rendering and a lack of detail when used for a Gaussian splatting framework.
Consequently, we densified the mesh to 84,317 points by recursively splitting edges and faces if the length or area exceeds pre-defined thresholds.
This allowed the 3DGS fitting to reach almost 30 PSNR on reconstruction in less than a minute on NVIDIA GeForce RTX 3090.

We trained our models with the AdamW~\cite{Loshchilov2017DecoupledWD} optimizer. 
For the training of the 3D reconstruction model, we utilize LPIPS~\cite{lpips} (0.15), L1 (1.0), and SSIM (0.25) losses.
For the diffusion model, we used the ``\emph{pred\_x0}'' objective, number of timestamps 1000, and the L2 loss.


\begin{figure*}[t]
    \centering
    \includegraphics[width=0.99\linewidth]{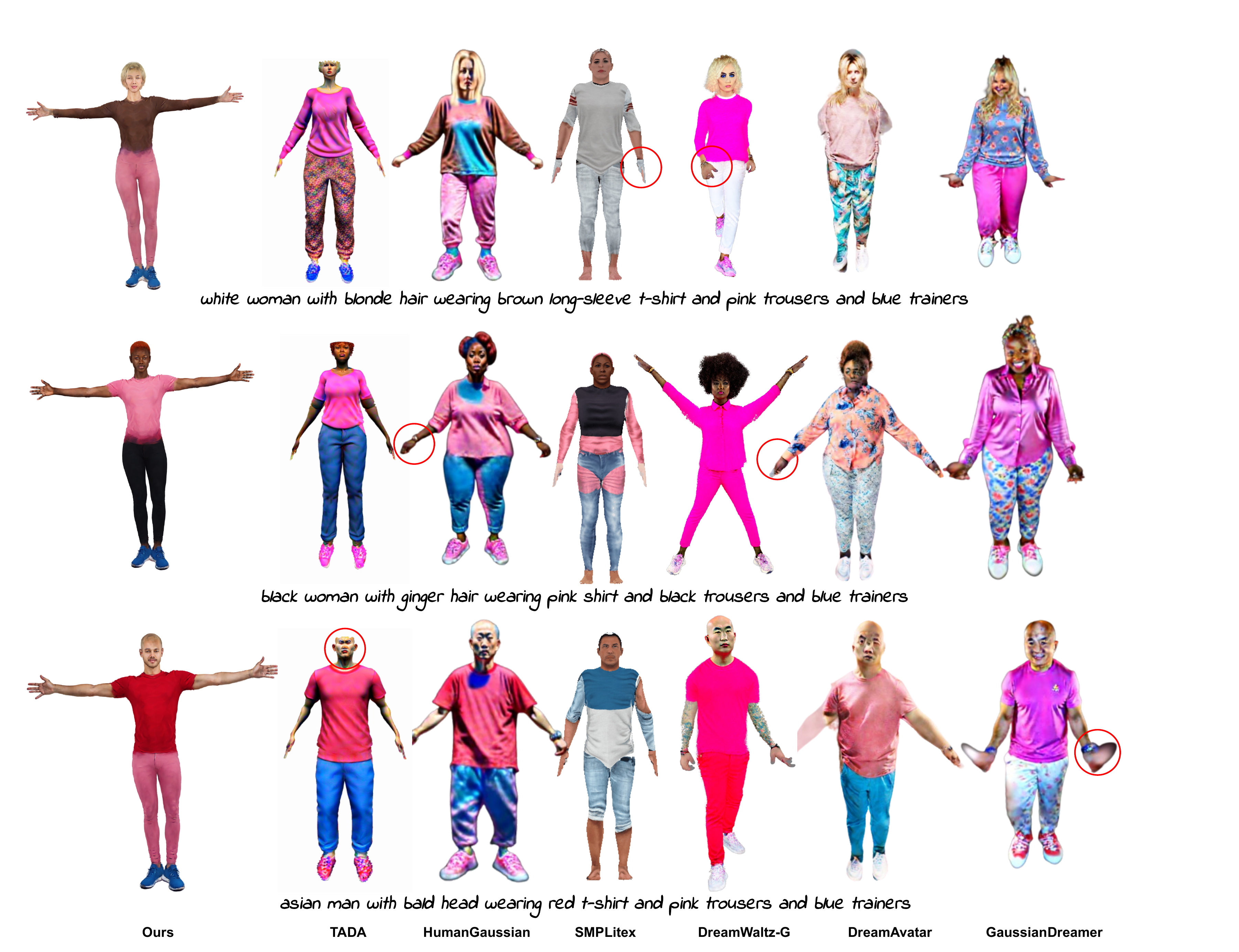}
    \vspace{-0.4cm}
    \caption{Comparison of rendered images generated with text prompt conditioning. The rendering artifacts and problems on hands and faces are circled in red.}
    \label{fig:prompt_comparison}
\end{figure*}

\begin{figure*}[t]
    \centering
    \includegraphics[width=0.99\linewidth]{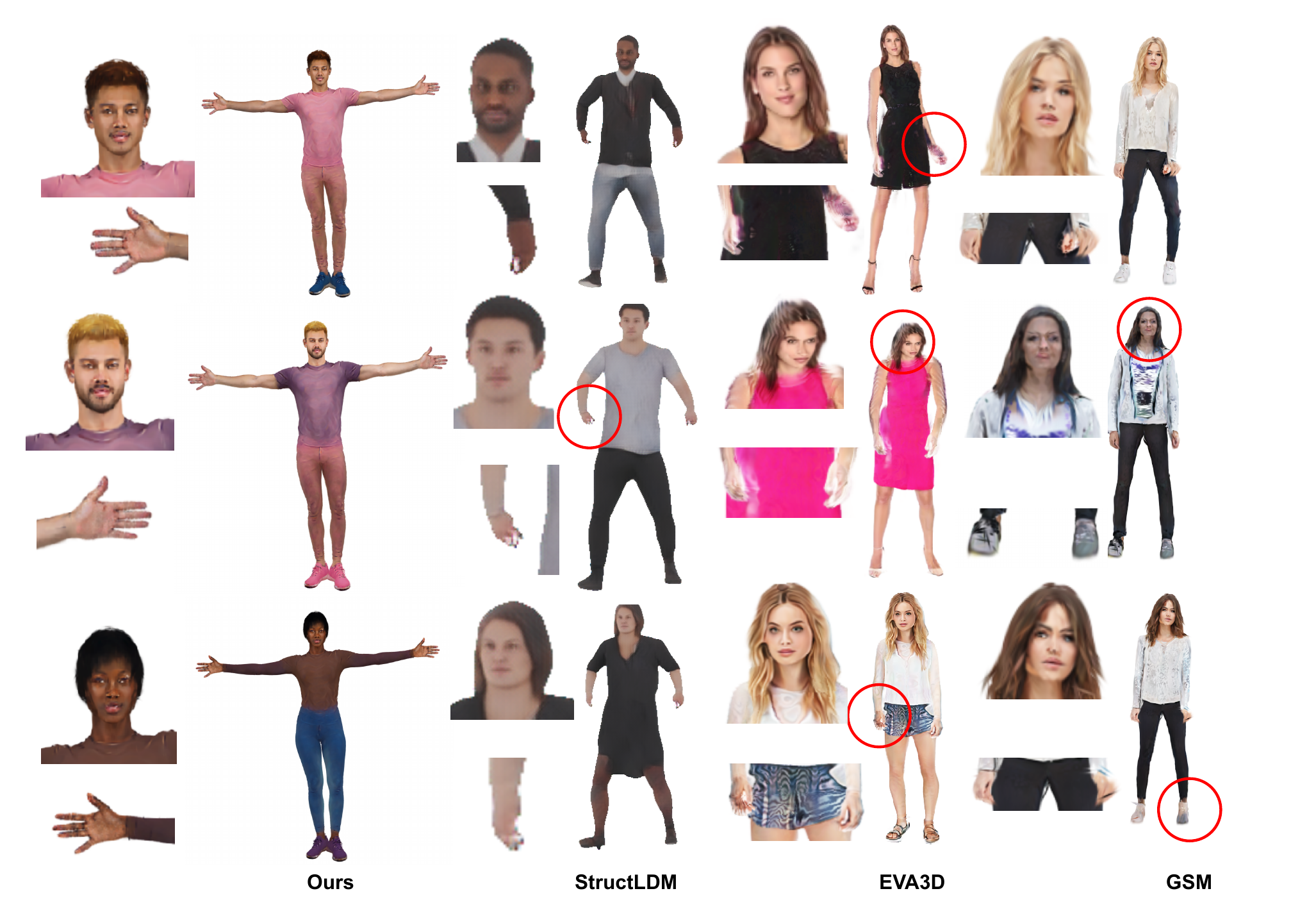}
    \vspace{-0.5cm}
    \caption{Comparison to images rendered by \emph{state-of-the-art} models that do not support text conditioning (we used random prompts for our method). Artifacts and issues, particularly on hands and faces, are circled in red.}
    \label{fig:uncond_comp}
\end{figure*}


The time comparison is shown in Table~\ref{tab:speed_comparison}.
Our method outperforms the text-conditioned ~\emph{state-of-the-art} models (sometimes by orders of magnitude) in terms of time to generate and render a human.
Text-conditioned methods such as HumanGaussian~\cite{liu2023humangaussian}, DreamWaltz-G~\cite{huang2024dreamwaltz-g} or DreamAvatar~\cite{cao2024dreamavatar} first generate training images, then fit a 3DGS to them, and then render (\eg, Instant-NGP~\cite{mueller2022instant} or NeRF~\cite{mildenhall2020nerf}).
With the exception of SMPLitex~\cite{casas2023smplitex}, no other text-guided model in the competitors list renders a 3D human directly.
The methods that unconditionally generate humans have a comparative speed, but they cannot align the generation with the text prompt.

\subsection{Quantitative Evaluation}
We quantitatively evaluate 3D human generation with large language models in Table~\ref{tab:score_image_eval}, focusing on prompt alignment and image aesthetics (including realism). While CLIP-based metrics~\cite{Radford2021LearningTV} have been adopted to assess the consistency between text prompts and generated images~\cite{liu2023humangaussian}, we found the CLIP score to be unreliable due to factors such as resolution differences and variations in human articulation. We rendered 10 images per method; however, the high computational cost of some approaches makes large-scale quantitative evaluation impractical.
Our method achieved the best results in image aesthetics and text prompt alignment, outperforming \emph{state-of-the-art} methods by a high margin, as evaluated by all four language models.
Almost 6 days of continuous GPU computing were needed to generate competitor results.
In our evaluation, we excluded some relevant works (SIGMAN~\cite{yang2025sigmanscaling3dhumangaussian}, Text2Avatar~\cite{gong2024text2avatar}, DreamHuman~\cite{DreamHuman}, and~\cite{Kolotouros_instant}) as the code is not publicly available.


\begin{figure*}[t]
    \centering
    \includegraphics[width=0.33\linewidth]{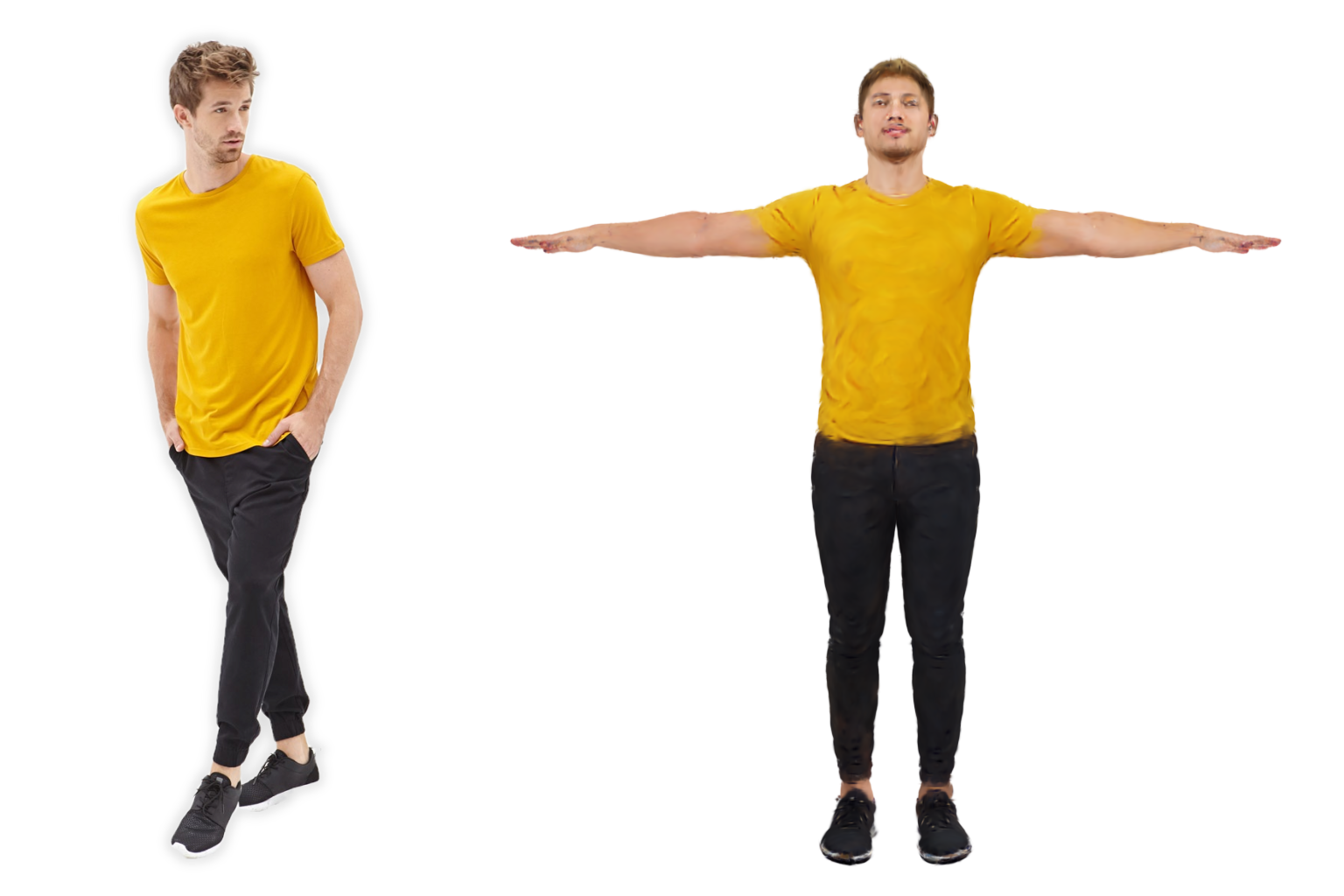}
    \includegraphics[width=0.2175\linewidth]{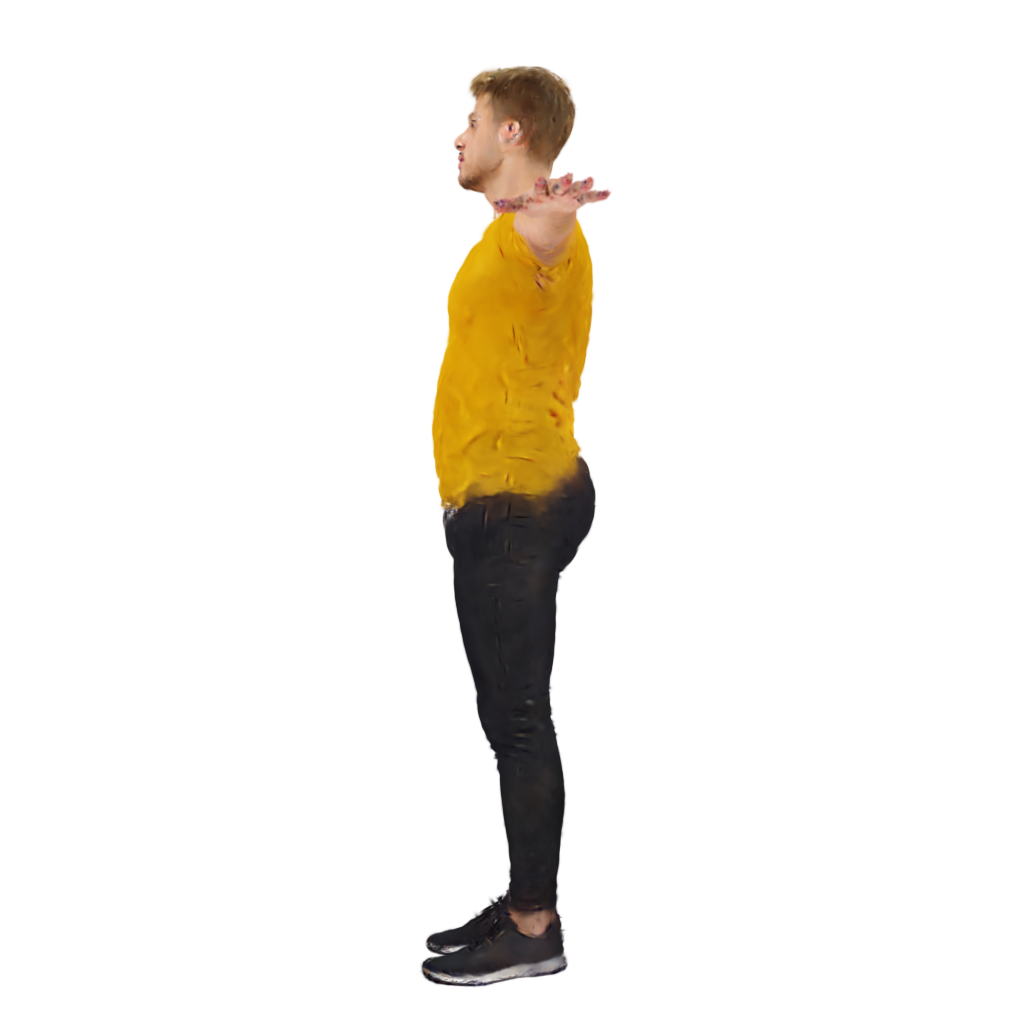}
    \includegraphics[width=0.2175\linewidth]{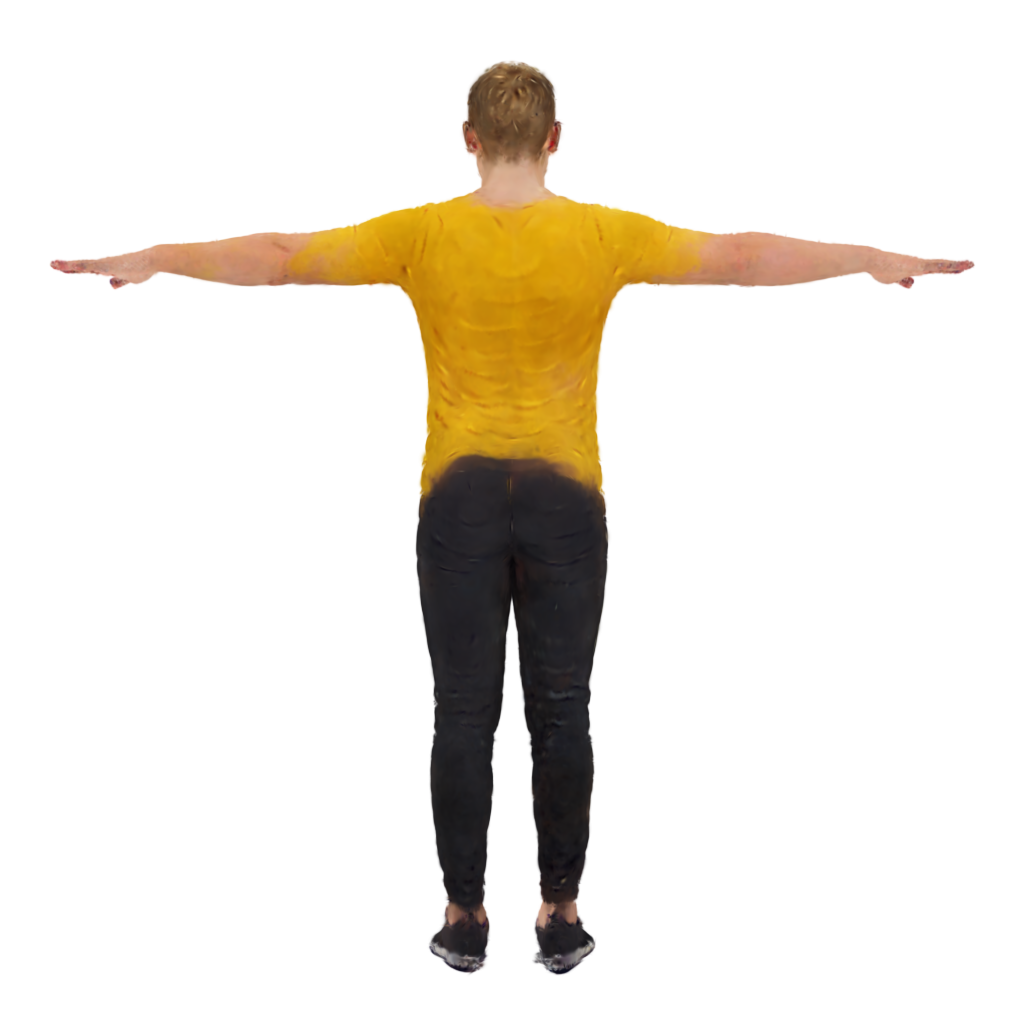}
    \includegraphics[width=0.2175\linewidth]{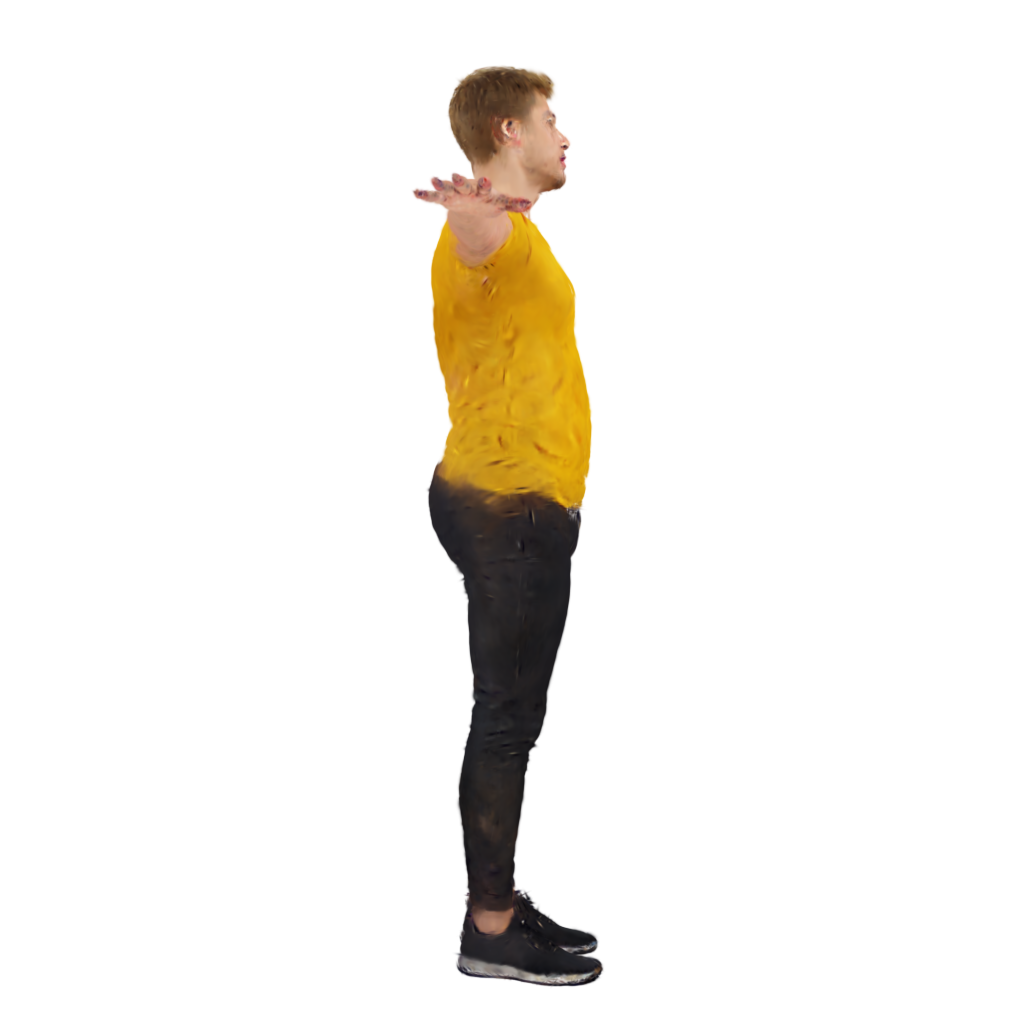}

    \vspace{-0.15cm}

    \caption{Novel view reconstruction on out-of-distribution image from the SHHQ~\cite{stylegan} dataset. The input image is on the left.}
    \label{fig:in-the-wild_main}
\end{figure*}

\begin{figure*}[t]
    \centering

    \includegraphics[width=0.24\linewidth]{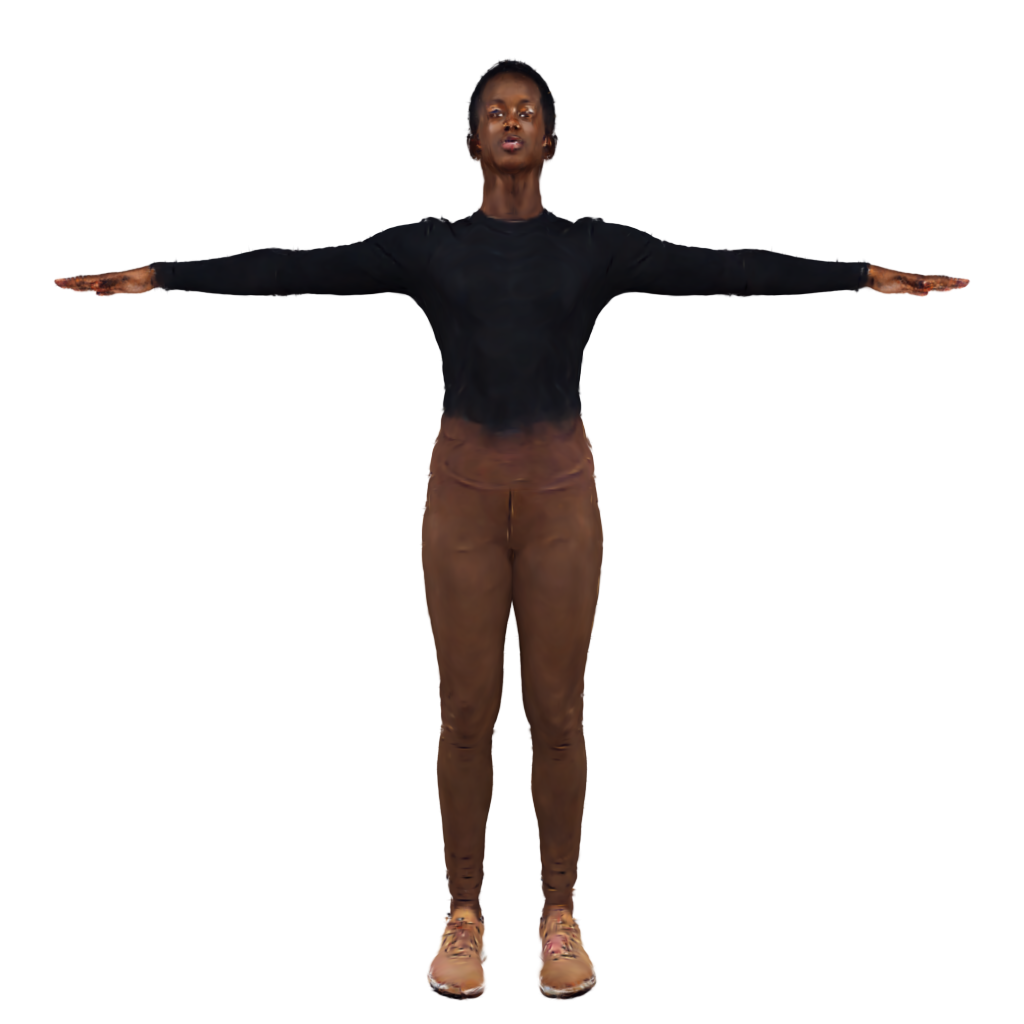}
    \includegraphics[width=0.24\linewidth]{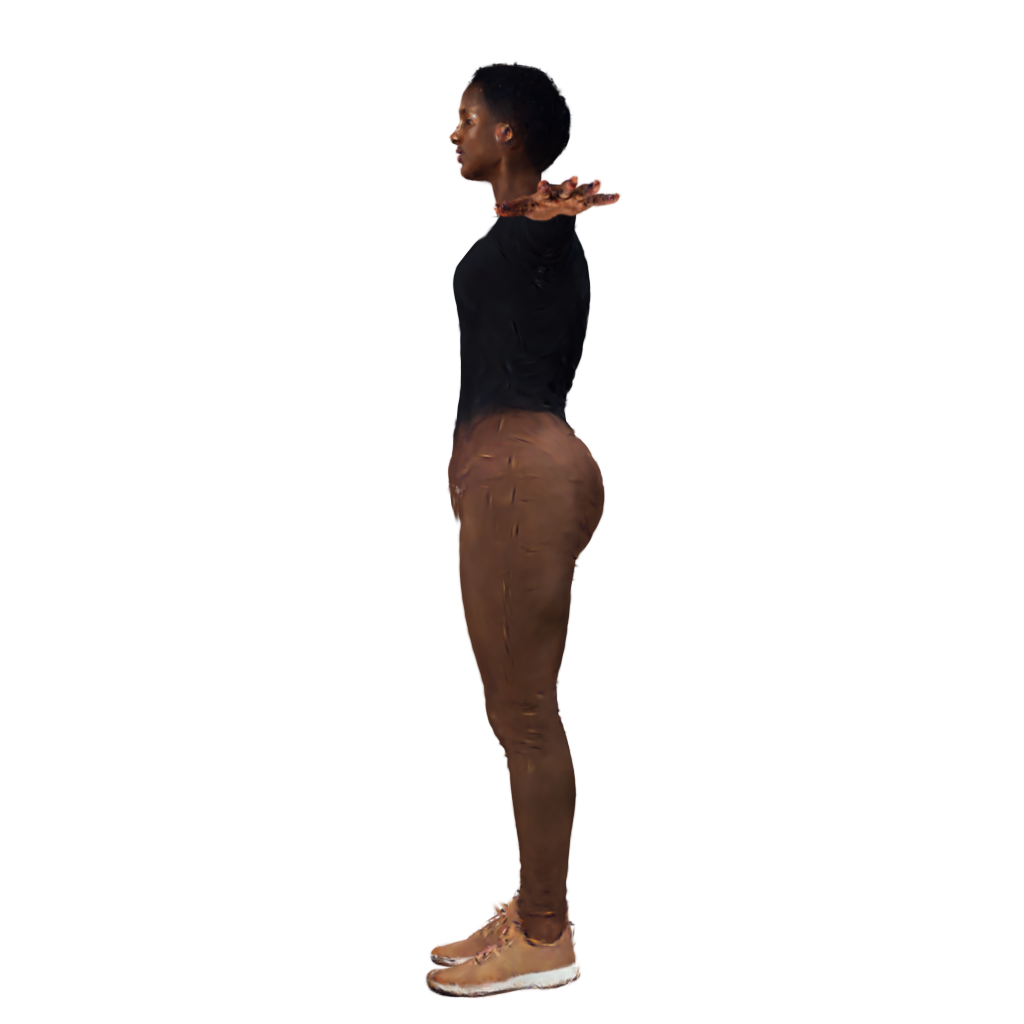}
    \includegraphics[width=0.24\linewidth]{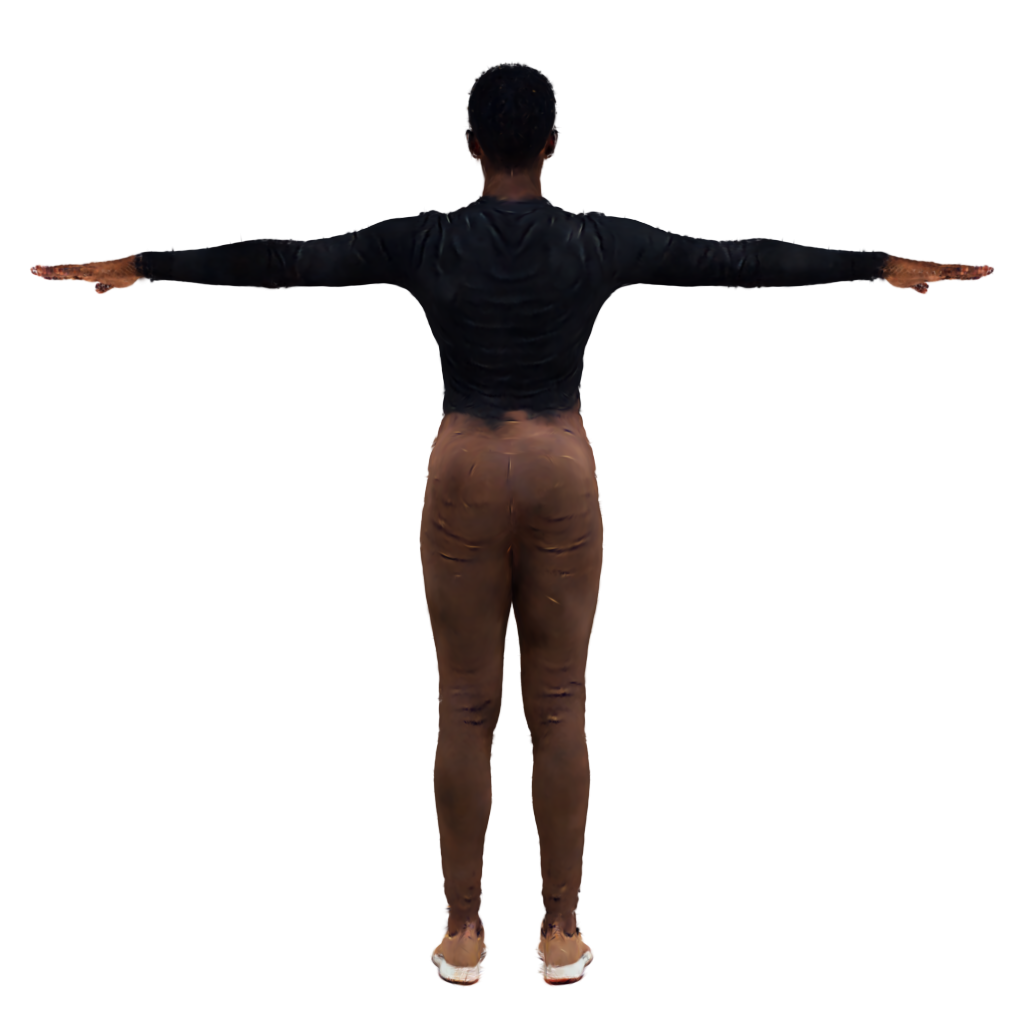}
    \includegraphics[width=0.24\linewidth]{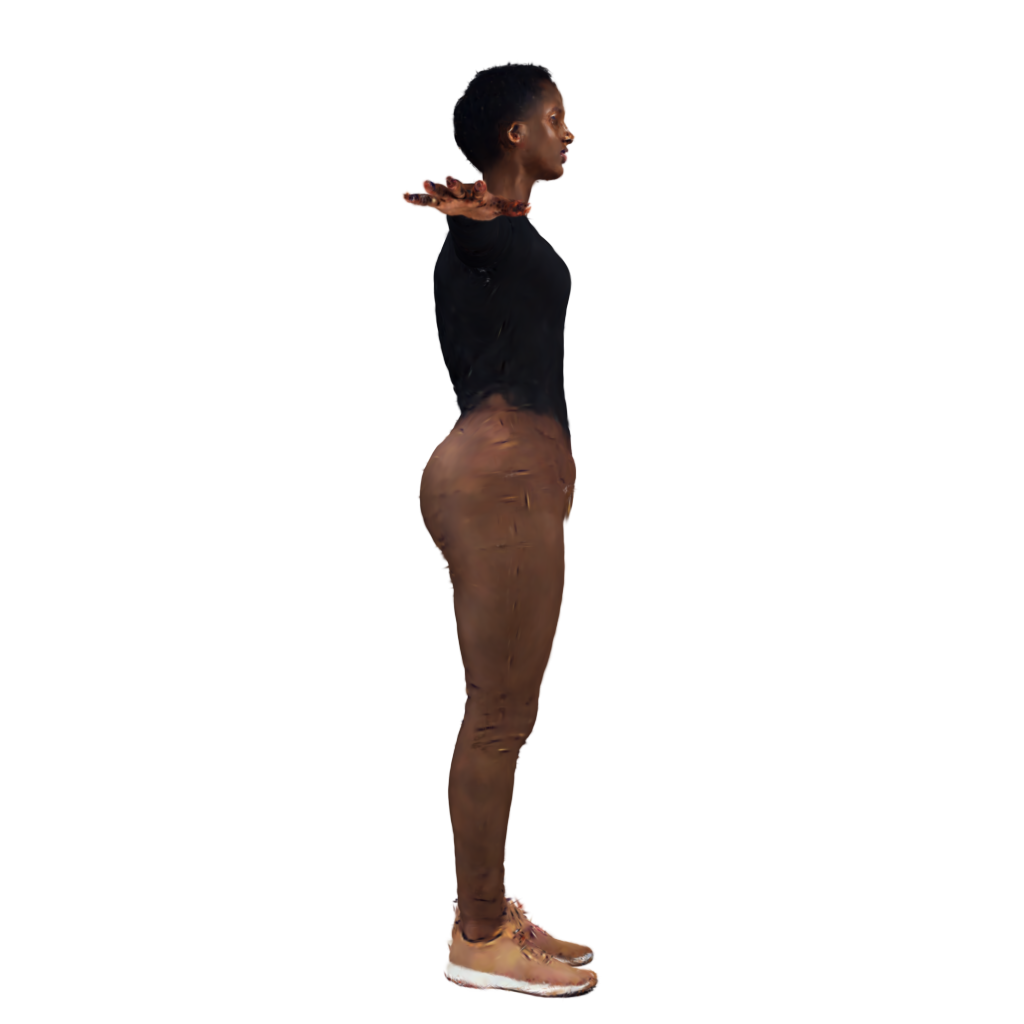}    
    \vspace{-0.15cm}
    \caption{Examples of novel view rendering of a 3D human generated via the proposed diffusion model.}
    \label{fig:diff_novel_views_main}
\end{figure*}

\begin{figure*}[h]
    \centering
    \includegraphics[width=0.24\linewidth]{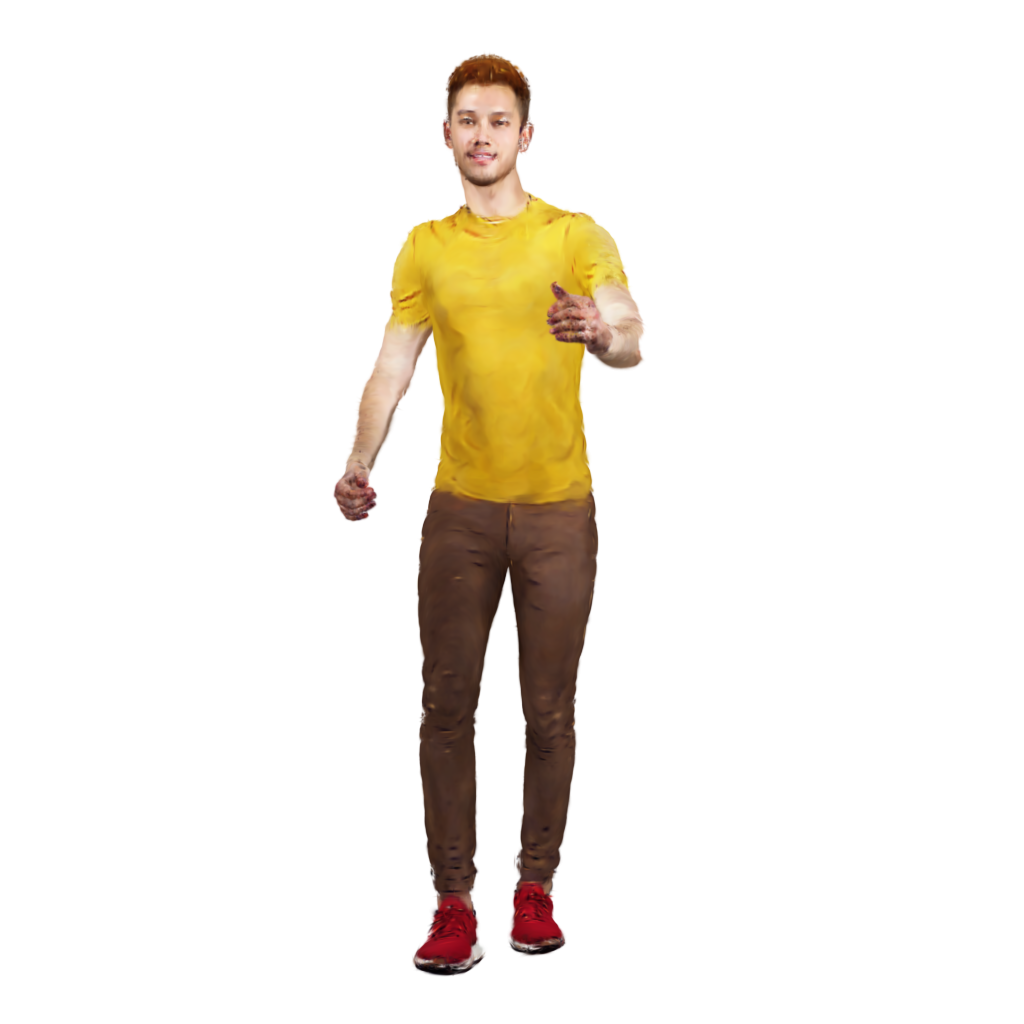}
    \includegraphics[width=0.24\linewidth]{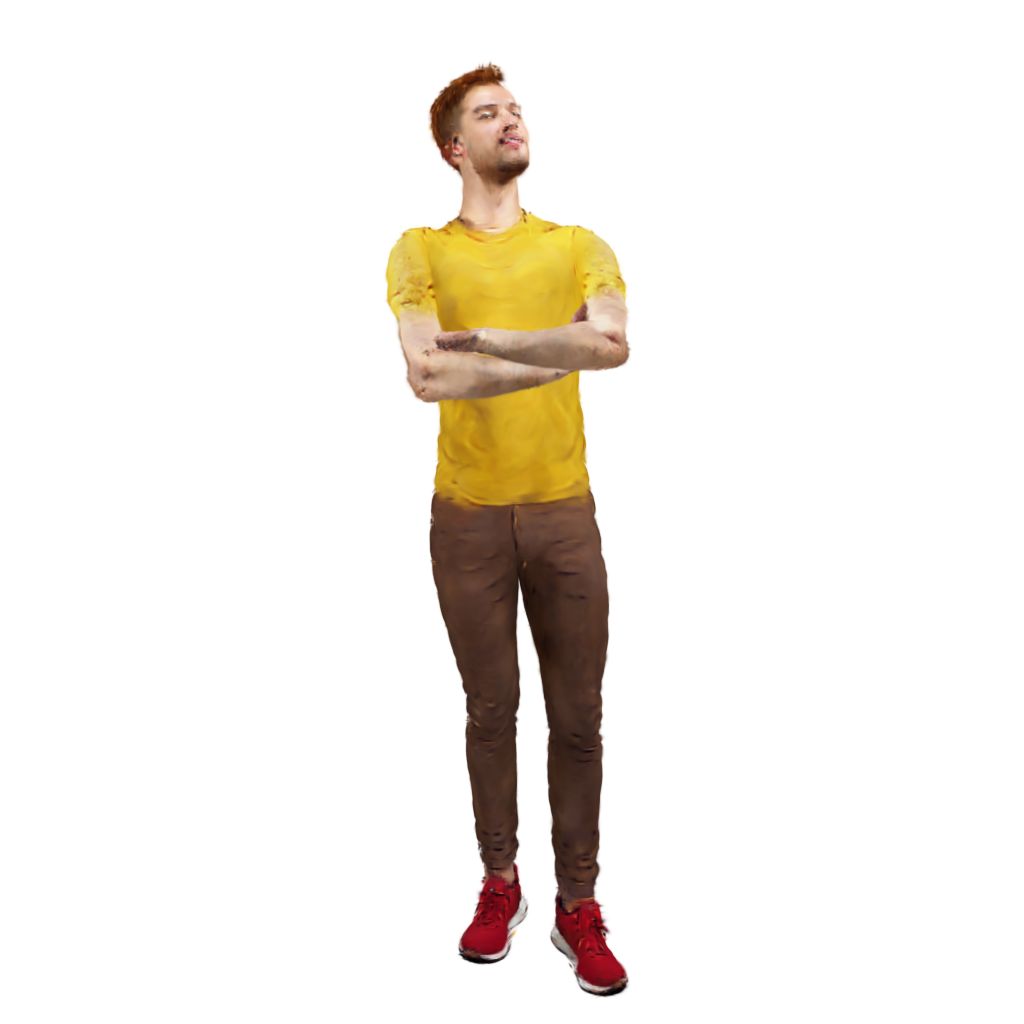}
    \includegraphics[width=0.24\linewidth]{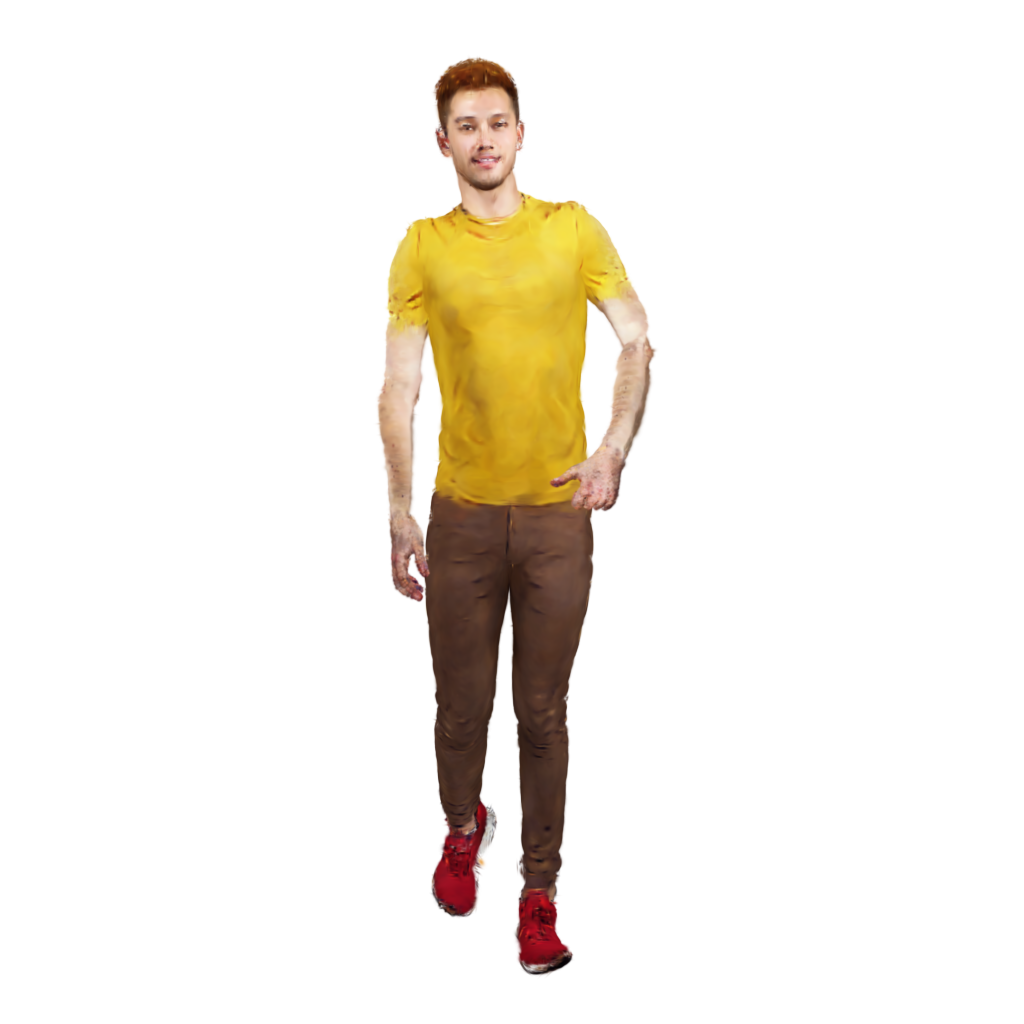}
    \includegraphics[width=0.24\linewidth]{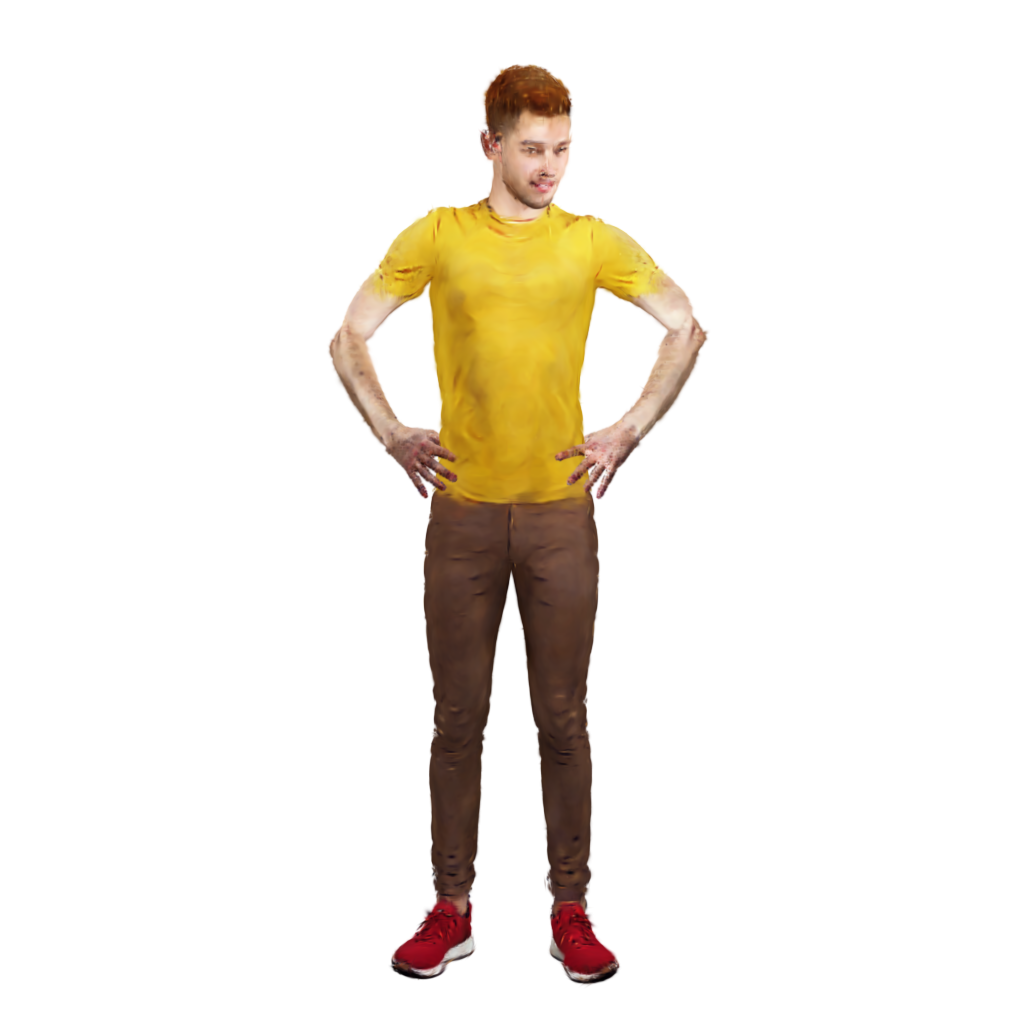}
    \vspace{-0.15cm}
    \caption{A generated human via our diffusion model rendered in various SMPL-X articulation poses from the AGORA~\cite{Patel:CVPR:2021} dataset.}

    \label{fig:animation}
\end{figure*}

\subsection{Qualitative Evaluation}
Fig.~\ref{fig:prompt_comparison} shows a qualitative comparison with text-guided~\emph{state-of-the-art} methods.
None of the competitors match the prompt we provided.
The reason for this could be that most of the methods use Stable Diffusion (SD)~\cite{Rombach_diff} or ControlNet~\cite{zhang2023adding} internally to generate training images, and they struggle with prompt guidance.
Furthermore, the generated appearances look rather cartoonish, human proportions are strange, and the hand quality is poor.

Fig.~\ref{fig:uncond_comp} illustrates a visual comparison to the~\emph{state-of-the-art} models that generate without control of human appearance.
The reason for this is that these methods were trained primarily on real-world fashion datasets (\emph{e.g.}, SHHQ~\cite{stylegan}, DeepFashion~\cite{liuLQWTcvpr16DeepFashion}, UBCFashion~\cite{DBLP:conf/bmvc/ZablotskaiaSZS19}) that may have too diverse clothing to annotate with text.
Furthermore, those datasets could contain a bias towards female participants, because methods such as EVA3D~\cite{hong2023evad} and GSM~\cite{gsm} tend to generate women.
In Fig.~\ref{fig:uncond_comp}, we highlight the rendering artifacts that occurred on hands, feet, and faces, where the~\emph{state-of-the-art} methods struggle the most.
Our approach has fewer rendering artifacts, and the rendering quality looks higher.


Leveraging the diversity of training images, featuring a wide range of human appearances and viewing angles, our models generalize well to out-of-distribution human images and novel viewpoints.
First, Fig.~\ref{fig:in-the-wild_main} showcases the quality of multi-view reconstruction on an image from the SHHQ dataset, which was not included in the training set.
While our model successfully reconstructs the overall appearance, fine details such as facial features and clothing wrinkles are less accurate, likely due to domain differences in the data distribution.
Fig.~\ref{fig:diff_novel_views_main} demonstrates multi-view renderings of a 3D human generated by the proposed diffusion model.
The overall structure remains consistent across views, although slight blurring is visible along clothing edges.
Additionally, Fig.~\ref{fig:animation} shows renderings of the generated human in various SMPL-X articulation poses.
The person appears rigid, with the primary limitation being shading inconsistencies, as some body parts appear unnaturally bright.


%% file: tables/generation_speed_comparison.tex
\begin{table}[]
    \setlength{\tabcolsep}{5pt}
    \centering
    \begin{tabular}{l|r||l|r}
         {\bf Method} & {\bf Time}& {\bf Method} & {\bf Time}  \\ \hline
         
         HumanGauss~\cite{liu2023humangaussian} & 68.79 & DrmWaltz-G~\cite{huang2024dreamwaltz-g} & 201.95 \\
         
         TADA~\cite{liao2024tada} & 191.74 & DrmAvatar~\cite{cao2024dreamavatar} & 350.50 \\
         
         GaussDrmer~\cite{yi2023gaussiandreamer} & 11.66 & SMPLitex~\cite{casas2023smplitex} & 0.29 \\
         
         EVA3d$^*$~\cite{hong2023evad} & 0.16 & StructLDM$^*$~\cite{StructLDM} & 0.10 \\
         
         GSM$^*$~\cite{gsm} & 0.12 & Ours & 0.45 \\
    \end{tabular}
    \caption{Time comparison (in minutes) for one sample generation. The methods denoted with $^*$ do not have control over appearance.}
    \label{tab:speed_comparison}
\end{table}

%% file: tables/score_image_evaluation.tex
\begin{table*}[]
    \centering
    \setlength{\tabcolsep}{15pt}
    \begin{tabular}{l|c|c|c|c}

         \multirow{2}{*}{{\bf Method}} & Copilot & Gemini & Grok & Claude  \\

          & Aln. / Aes. & Aln. / Aes.  & Aln. / Aes. & Aln. / Aes.    \\ \hline

         TADA~\cite{liao2024tada} & 0.59 / 0.65& 0.70 / 0.24 & 0.76 / 0.68 & 0.76 / 0.63 \\
         HumanGaussian~\cite{liu2023humangaussian} & 0.63 / 0.59 & 0.79 / 0.49 & 0.71 / 0.53  & 0.84 / 0.74  \\
         DreamAvatar~\cite{cao2024dreamavatar} & 0.49 / 0.52 & 0.82 / 0.32 & 0.57 / 0.38 & 0.72 / 0.63 \\
         GaussianDreamer~\cite{yi2023gaussiandreamer} & 0.34 / 0.46  &  0.50 / 0.34 &  0.57 / 0.66 & 0.46 / 0.62  \\
         DreamWaltz-G~\cite{huang2024dreamwaltz-g,huang2024dreamwaltz} & 0.49 / 0.56 &  0.56 / 0.69 & 0.71 / 0.65 & 0.77 / 0.77   \\
         SMPLitex~\cite{casas2023smplitex} & 0.32 / 0.39 & 0.30 / 0.11 & 0.21 / 0.40 & 0.52 / 0.31   \\
         Ours & {\bf 0.79} / {\bf 0.68}  & {\bf 0.88} / {\bf 0.79} & {\bf 0.82} / {\bf 0.70} & {\bf 0.91} / {\bf 0.84}\\
    \end{tabular}
    \caption{Comparison evaluation of rendered images in terms of text-prompt alignment (``Aln.'') and image aesthetics (``Aes.'') using large language models. The values range from 0 (worst) to 1 (best).}
    \label{tab:score_image_eval}
\end{table*}

%% file: sections/limitations.tex
\section{Limitations}
One of our limitations is the lack of patterns in clothing.
However, this is a trade-off between text control and detail. With more detail, it becomes more confusing to model which human textures to generate. 
Additionally, women in our pipeline have short hair, because the current SMPL-X model does not have hair parameters, and the hair is created by displacements from the mesh.
Finally, our model is still limited by certain combinations of the text prompts discussed in the Supplementary Materials.

%% file: sections/conclusions.tex
\section{Conclusions}
We presented a weakly supervised pipeline, where we first generate a photorealistic image with an~\emph{off-the-shelf} diffusion model and extract initial reconstructions through 3D model fitting. 
Next, we train the reconstruction model with efficient image features to learn the mapping to 3D point cloud features.
The reconstructed 3DGS parameters from a single reconstruction are then used in training a point-cloud text-conditioned diffusion model with classifier-free guidance. 

With extensive qualitative and performance evaluation, we show that our method outperforms the current~\emph{state-of-the-art} approaches in 3D human generation regarding rendered quality, speed, and appearance control.
Our images present high-fidelity textures, realistic proportions, with hand and face quality.
We plan to release the synthetic dataset alongside the 3D reconstruction and HuGeDiff code.


%% file: sections/acknowledgements.tex
\section{Acknowledgements}
This work was supported by the SNSF project `SMILE II' (CRSII5 193686), the Innosuisse IICT Flagship (PFFS-21-47), EPSRC grant APP24554 (SignGPT-EP/Z535370/1) and through funding from Google.org via the AI for Global Goals scheme. This work reflects only the author’s views and the funders are not responsible for any use that may be made of the information it contains.

%% file: sections/suppl.tex
\twocolumn[
\begin{center}
{\Large \textbf{Supplementary Material: HuGeDiff: 3D Human Generation via Diffusion with Gaussian Splatting}}
\end{center}
]

\begin{figure}[t]
    \centering
    \includegraphics[width=0.32\linewidth]{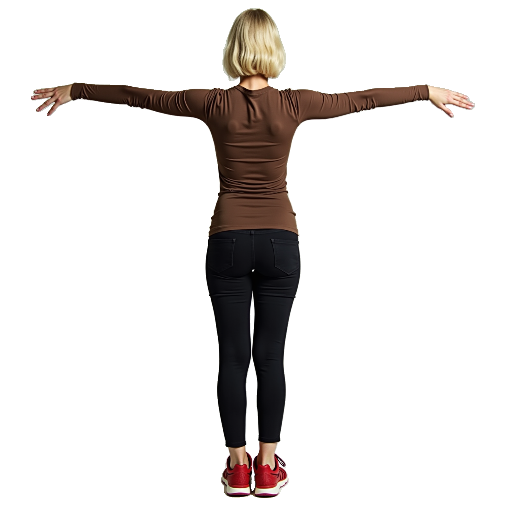}
    \includegraphics[width=0.32\linewidth]{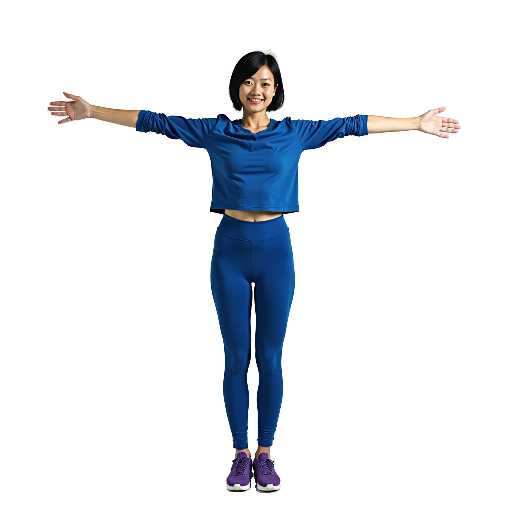}
    \includegraphics[width=0.32\linewidth]{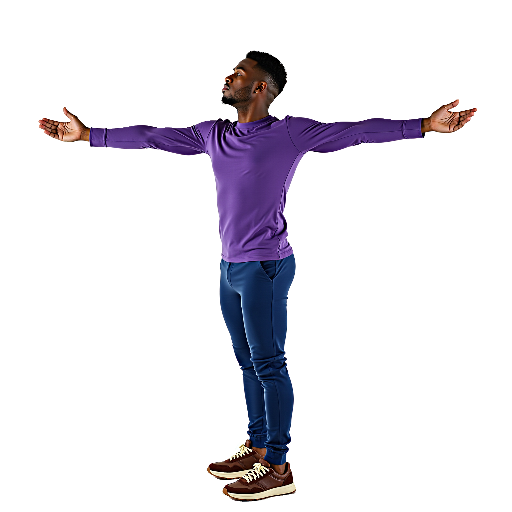}

    \includegraphics[width=0.32\linewidth]{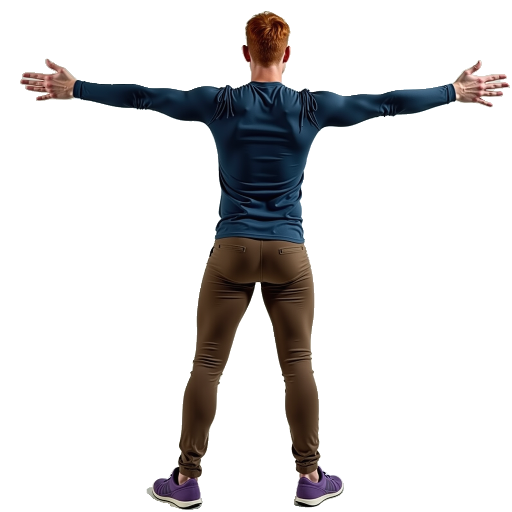}
    \includegraphics[width=0.32\linewidth]{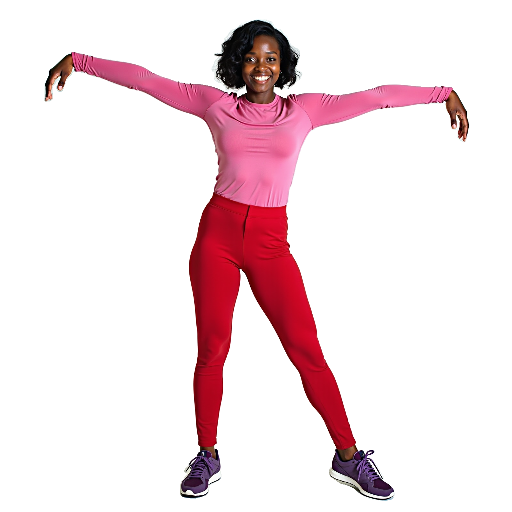}
    \includegraphics[width=0.32\linewidth]{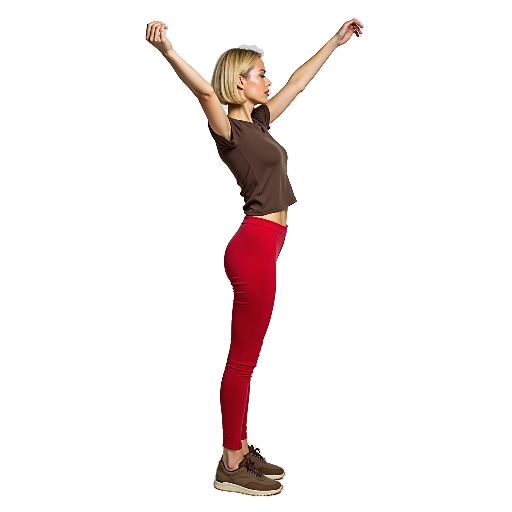}
    \caption{Examples of images generated by the FLUX diffusion model.}
    \label{fig:dataset}
\end{figure}
\begin{figure}[t]
    \centering
    \includegraphics[width=0.48\linewidth]{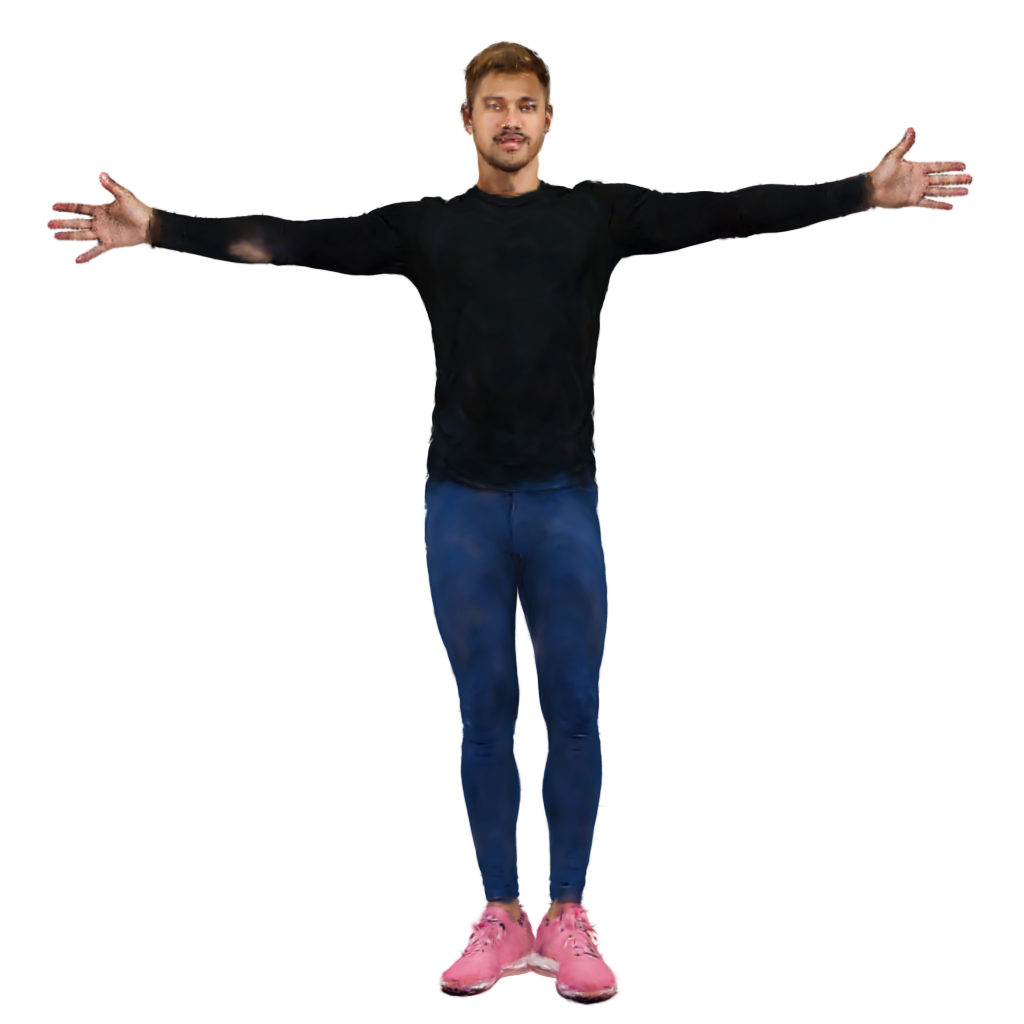}
    \includegraphics[width=0.48\linewidth]{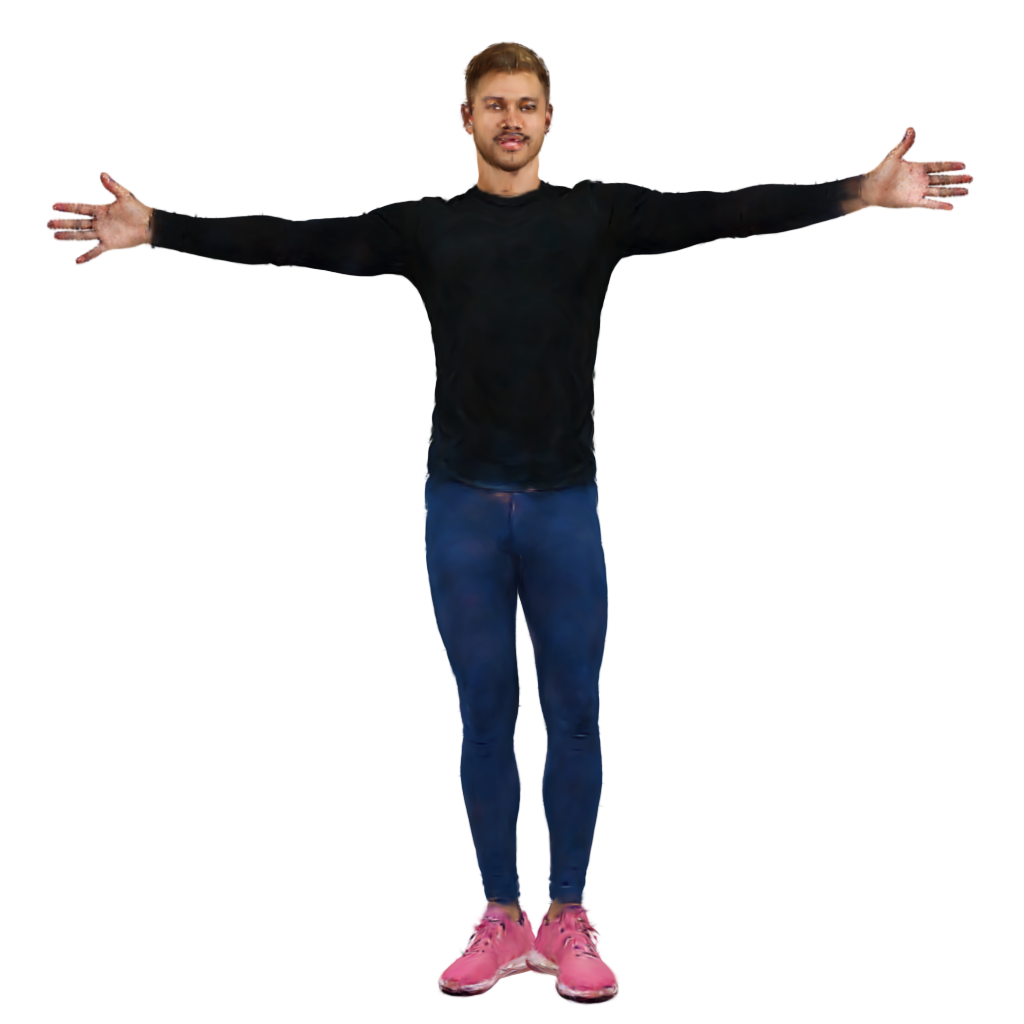}
    \includegraphics[width=0.48\linewidth]{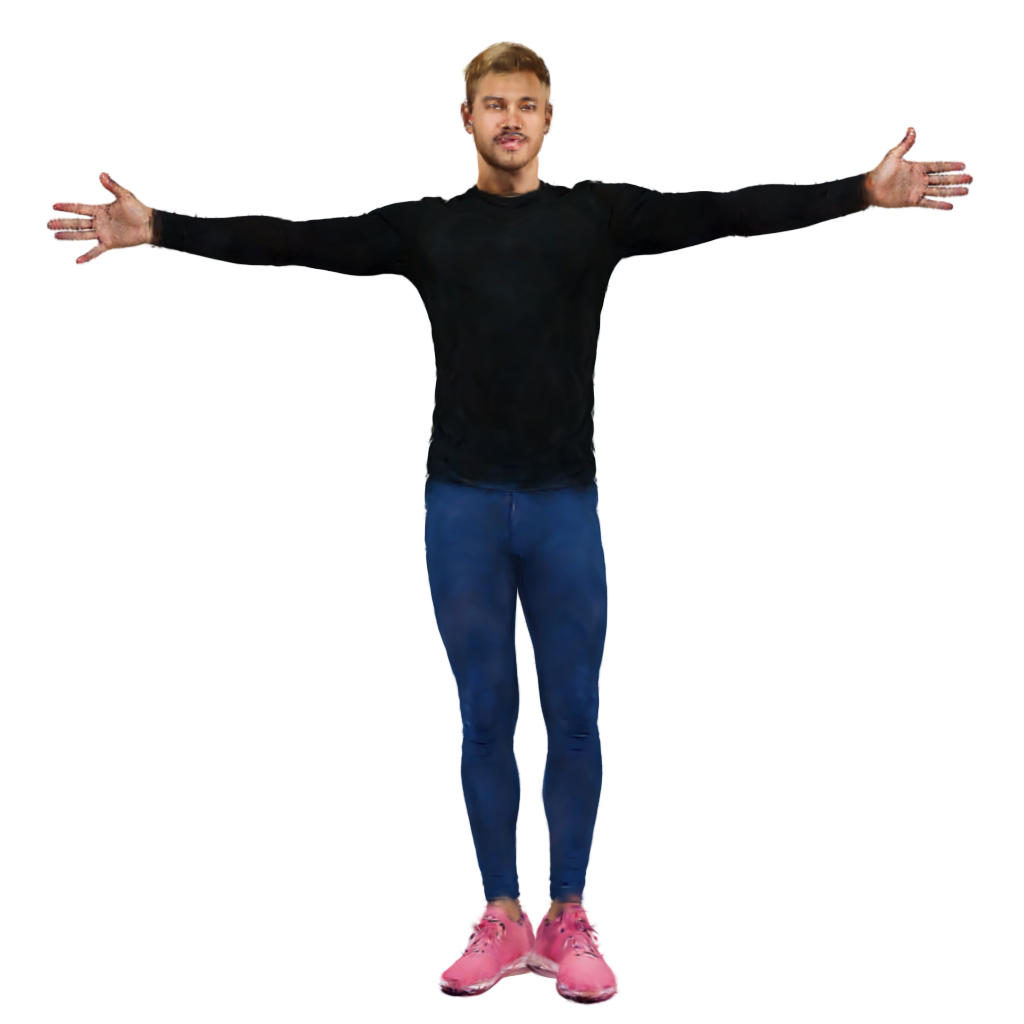}
    \includegraphics[width=0.48\linewidth]{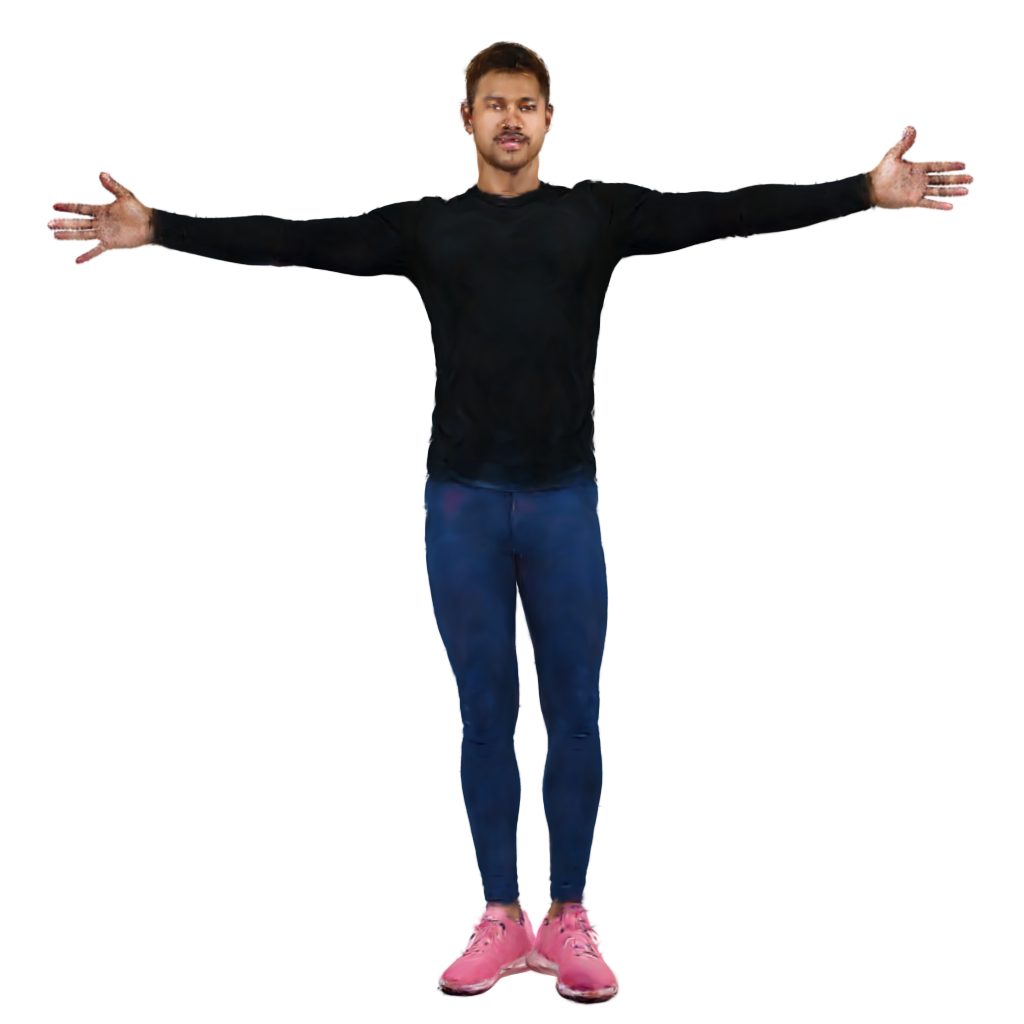}
    \caption{Same prompt with different seed diffusion sampling shows slight difference and variation in each rendered image.}
    \label{fig:same_prompt}
\end{figure}

\begin{figure}[t]
    \centering
    \includegraphics[width=0.48\linewidth]{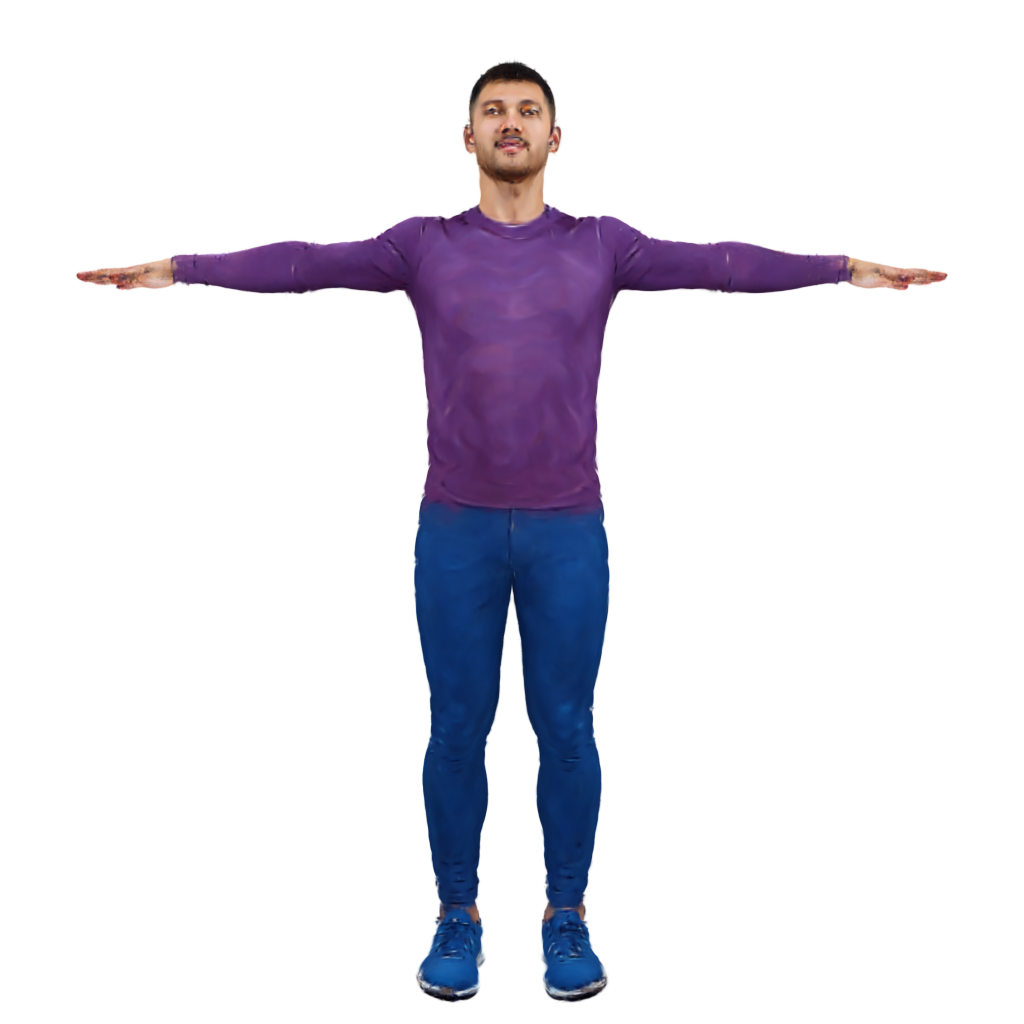}
    \includegraphics[width=0.48\linewidth]{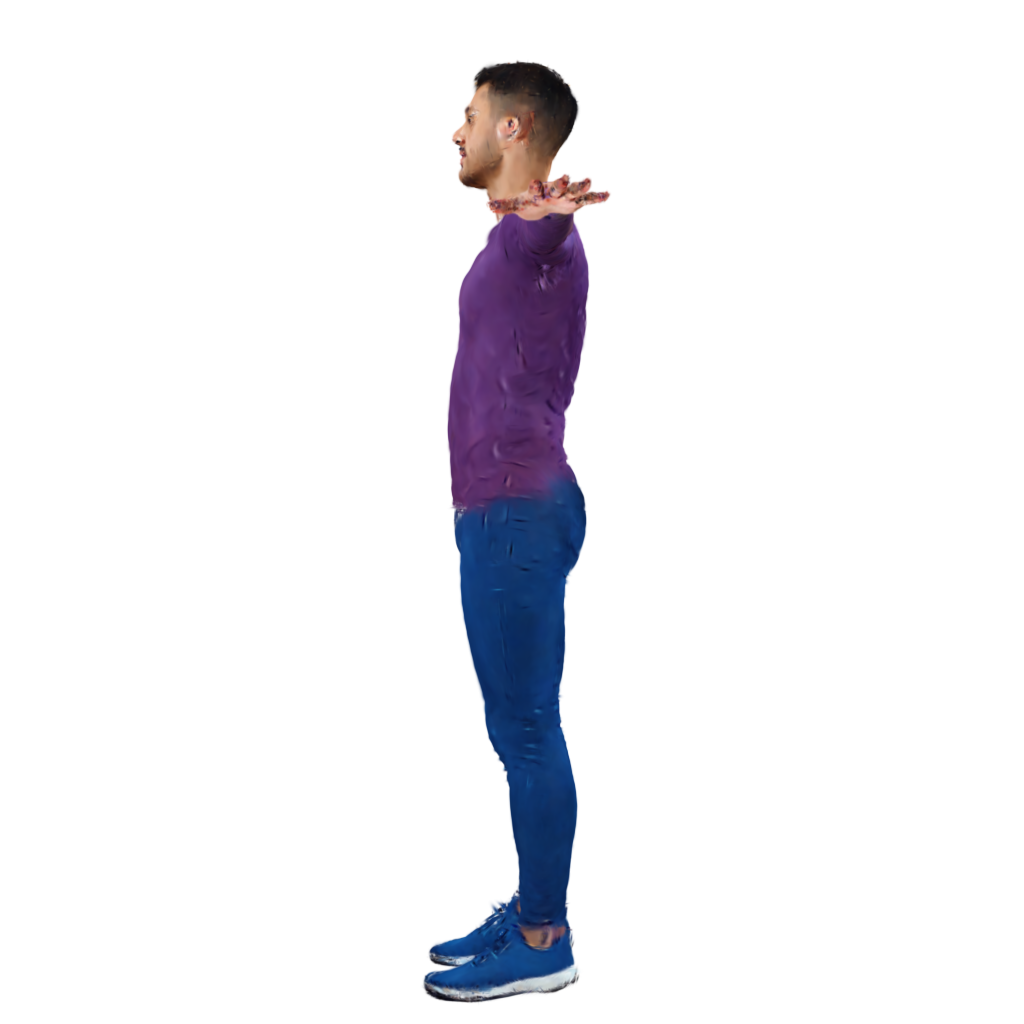}
    \includegraphics[width=0.48\linewidth]{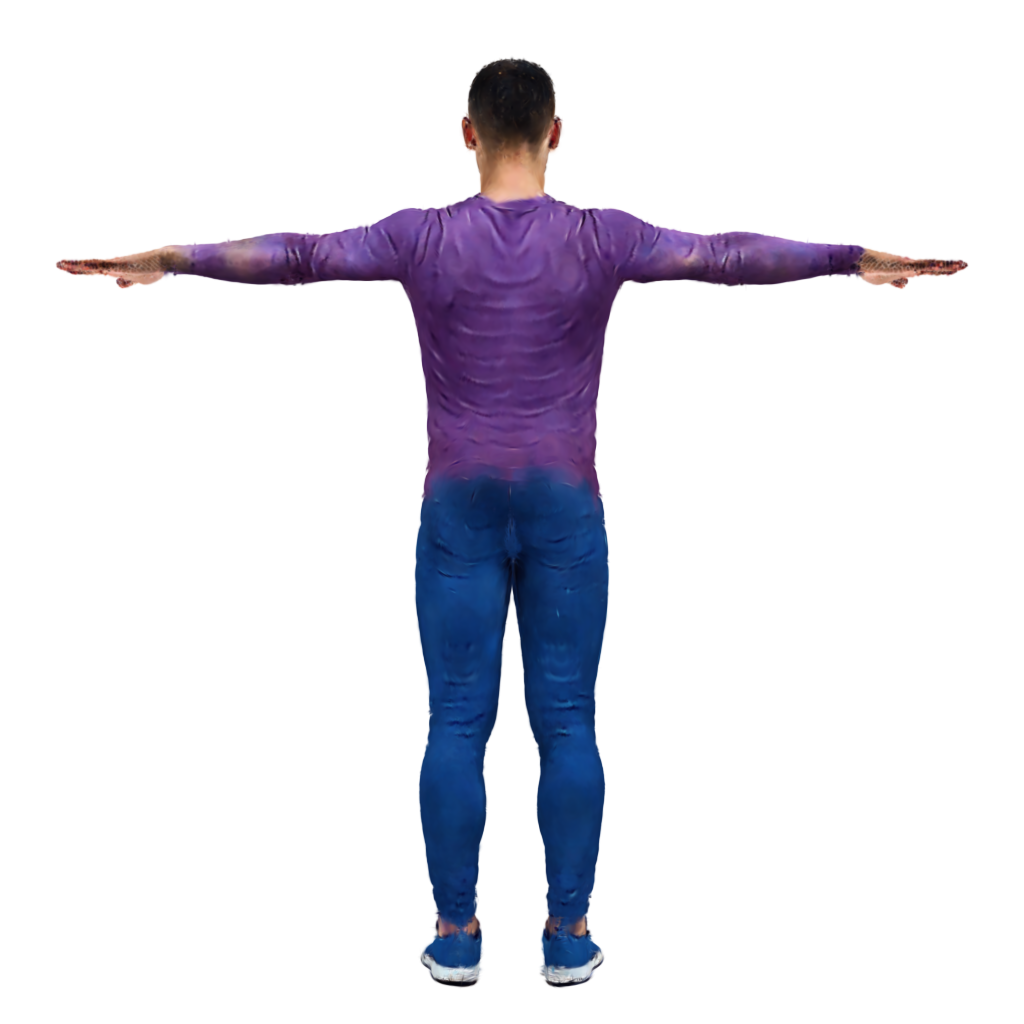}
    \includegraphics[width=0.48\linewidth]{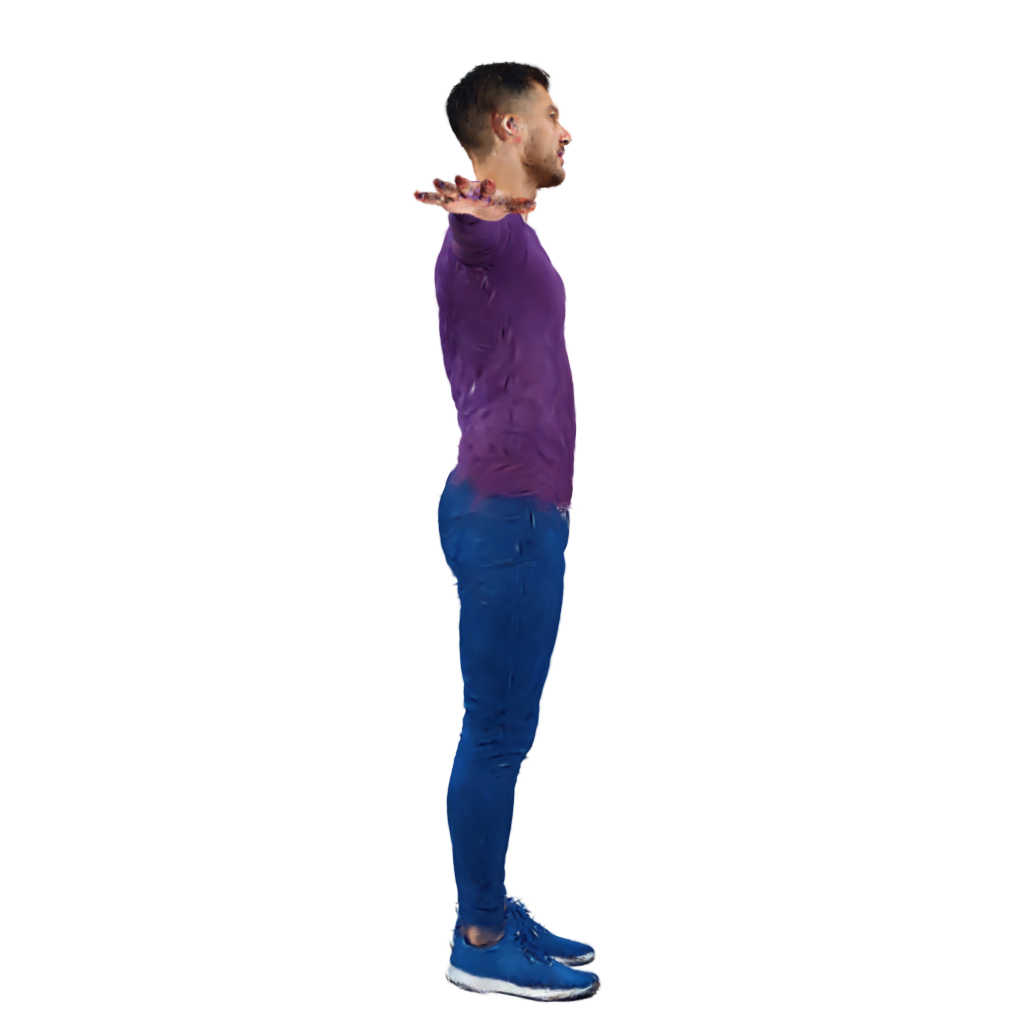}

    \caption{Novel view rendering of 3D humans generated by the diffusion model.}
    \label{fig:diff_novel_views}
\end{figure}

\section{Description of Equations}
Decoupling of image features to the point cloud. With query $(\mathbf{Q})$ being the point cloud of size $N$, and $M$ image features with dimension $C$ are both keys $(\mathbf{K})$ and values $(\mathbf{V})$, $h$ is number of heads:
\begin{equation}
    \mathbf{F}_X = \text{softmax}(\mathbf{Q} \, \mathbf{K}^\top) \, \mathbf{V},
\end{equation}
\begin{equation}
  \centering
  \begin{aligned}
    (\mathbb{R}^{N\times h \times 1 \times d } \, \mathbb{R}^{1\times h \times d \times M }) \, \mathbb{R}^{1\times h \times M \times \frac{C}{h}} \rightarrow \\ 
    \rightarrow \mathbb{R}^{N\times h \times 1 \times M } \, \mathbb{R}^{1\times h \times M \times \frac{C}{h}} \rightarrow \\ 
    \rightarrow  \mathbb{R}^{N\times h \times 1 \times \frac{C}{h}} \rightarrow  \mathbb{R}^{N\times C},
\end{aligned}
\end{equation}

Upsampling of point features from a smaller to a larger point cloud. The point-cloud features $l_\text{PE}(.)$ have dimension $d$, and the features matrix $\mathbf{F}_{X^\prime_0}$ has dimension $f$. $K \in \{1,\dots n\}^{N\times k}$ is an index matrix.
\begin{equation}
    \mathbf{F}_{X_0} \leftarrow \text{softmax}\big(l_\text{PE}(\tilde{\mathbf{X}}_0)l_\text{PE}(\tilde{\mathbf{X}}^\prime_0[K])^\top\big) \: \mathbf{F}_{X^\prime_0}[K].
\end{equation}
\begin{equation}
  \centering
  \begin{aligned}
    (\mathbb{R}^{N\times h \times 1 \times d } \, \mathbb{R}^{N\times h \times d \times k }) \, \mathbb{R}^{N\times h \times k \times \frac{f}{h}} \rightarrow \\ 
    \rightarrow \mathbb{R}^{N\times h \times 1 \times k } \, \mathbb{R}^{N\times h \times k \times \frac{f}{h}} \rightarrow \\ 
    \rightarrow  \mathbb{R}^{N\times h \times 1 \times \frac{f}{h}} \rightarrow  \mathbb{R}^{N\times f},
\end{aligned}
\end{equation}

For self-attention on point clouds, the keys $(\mathbf{K})$ include both point cloud features and relative 3D distances:
\begin{equation}
    \mathbf{K} = l_\text{PE}(\mathbf{F}_X[K^\prime]) + l_\text{PE}(\tilde{\mathbf{X}}_0 - \tilde{\mathbf{X}}_0[K^\prime])
\end{equation}

\begin{equation}
  \centering
  \begin{aligned}
    \mathbb{R}^{N\times h \times k \times \frac{d}{h} } + (\mathbb{R}^{N\times h \times 1 \times \frac{d}{h} } - \mathbb{R}^{N\times h \times k \times \frac{d}{h}}) \rightarrow \\ 
    \rightarrow \mathbb{R}^{N\times h \times k \times 
\frac{d}{h}}
    \end{aligned}
\end{equation}

\begin{figure*}[t]
    \centering
    \includegraphics[width=0.31\linewidth]{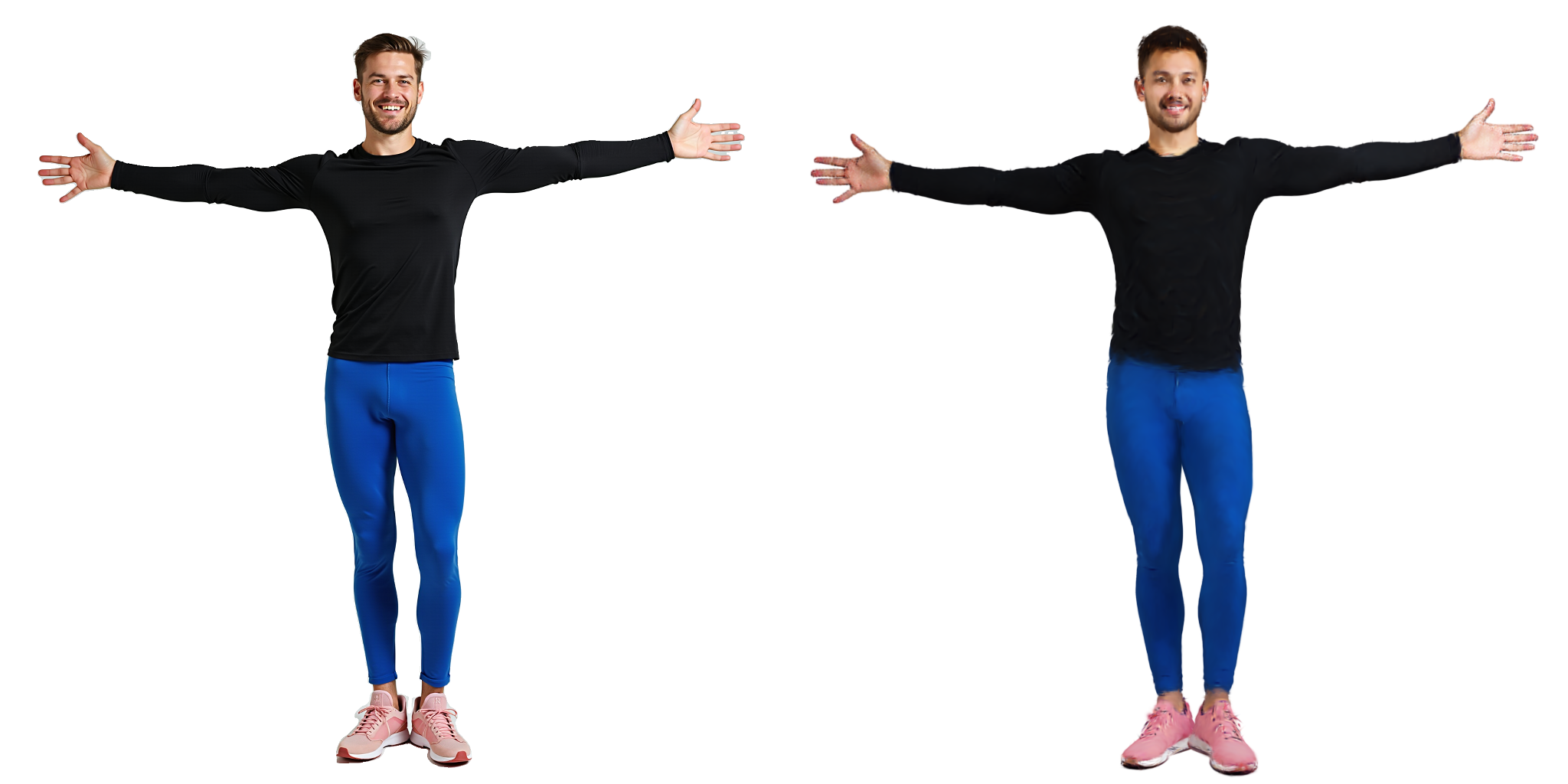}
    \includegraphics[width=0.16\linewidth]{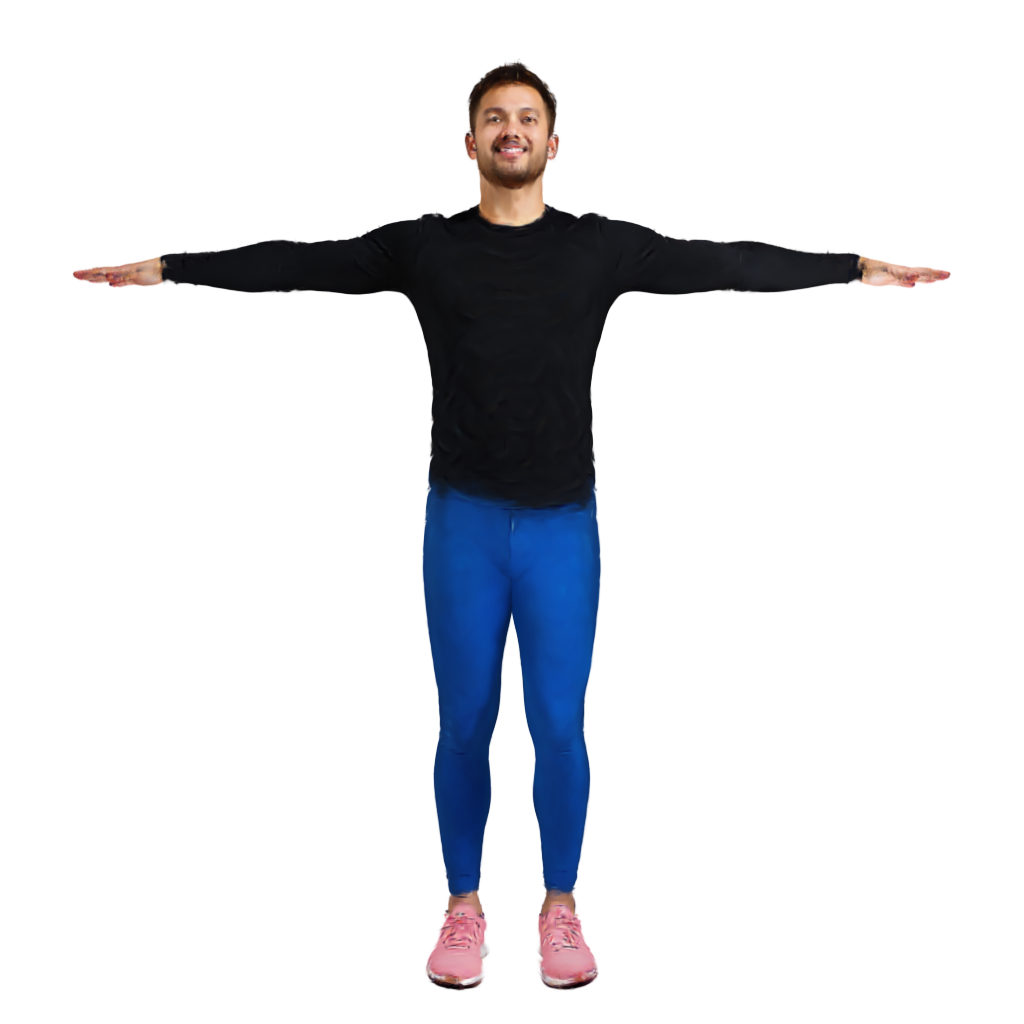}
    \includegraphics[width=0.16\linewidth]{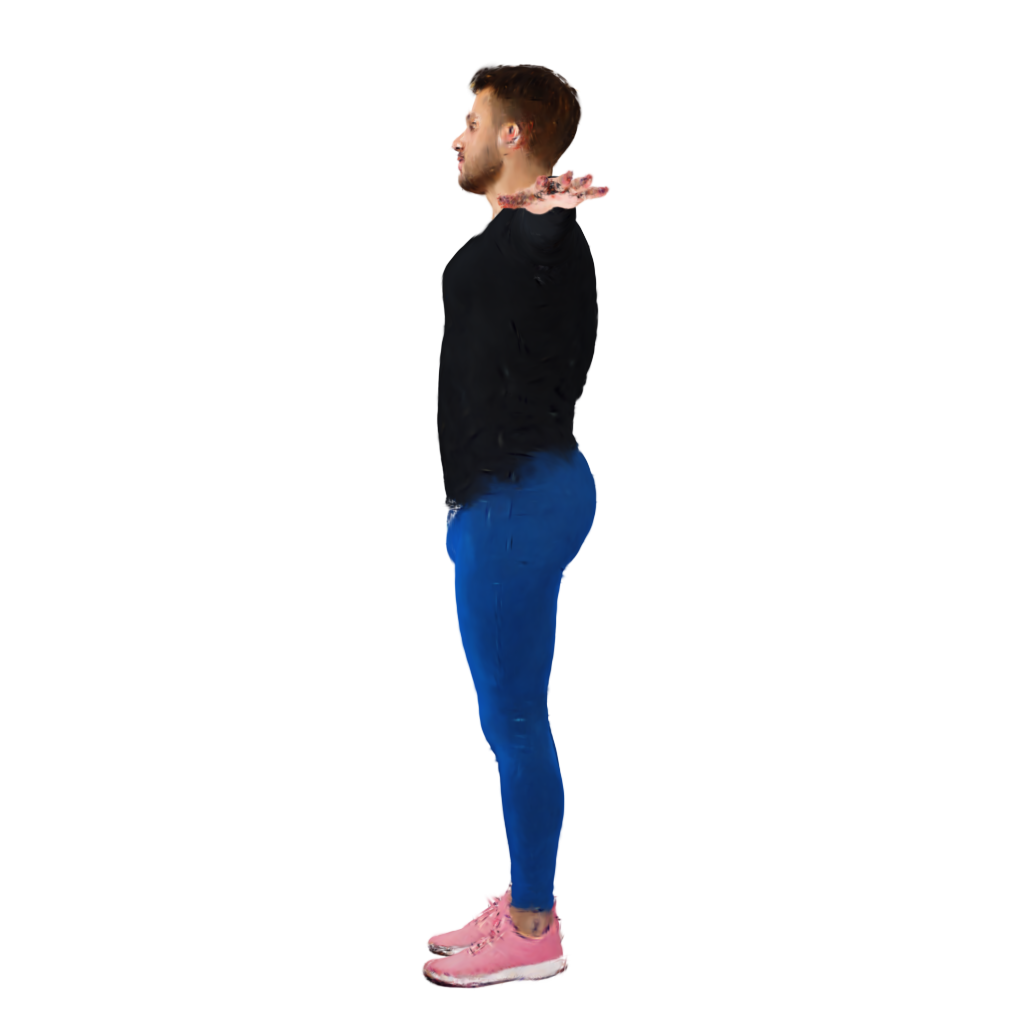}
    \includegraphics[width=0.16\linewidth]{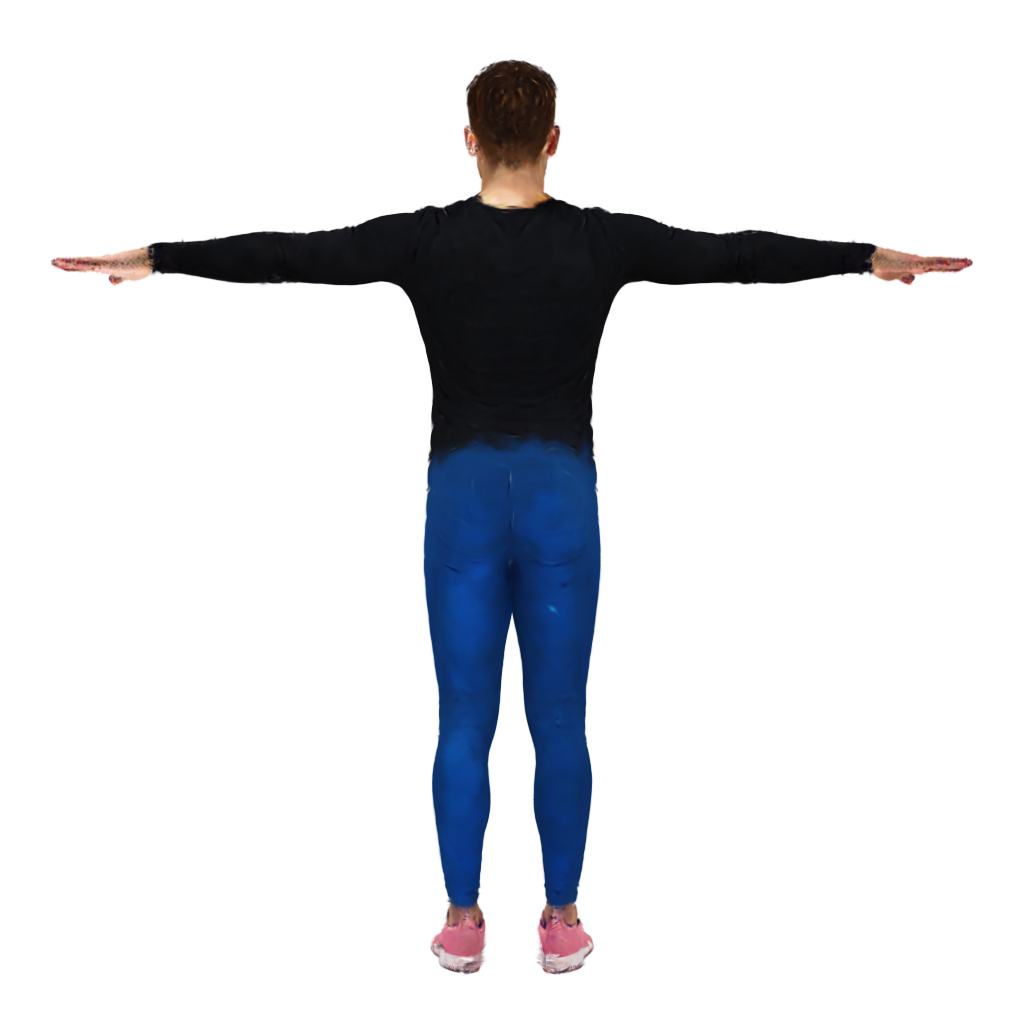}
    \includegraphics[width=0.16\linewidth]{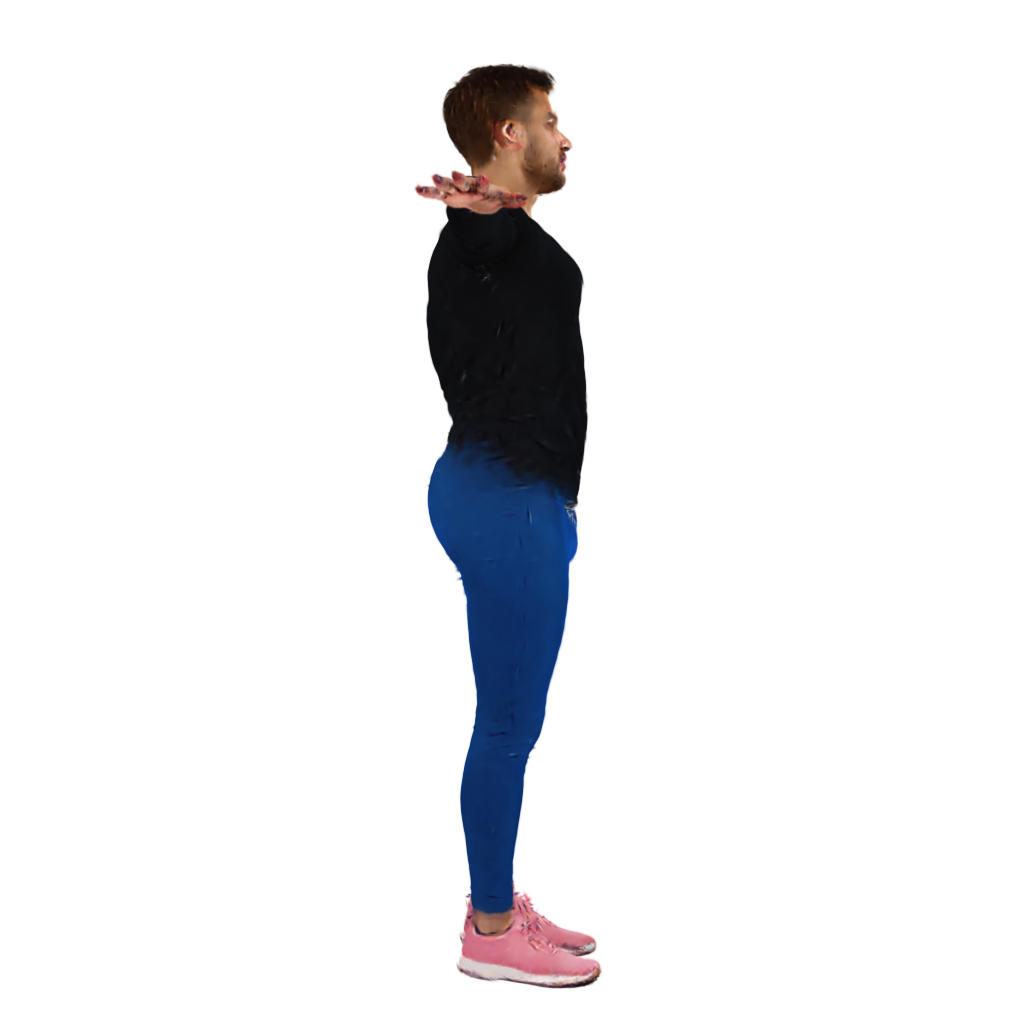}

    \includegraphics[width=0.31\linewidth]{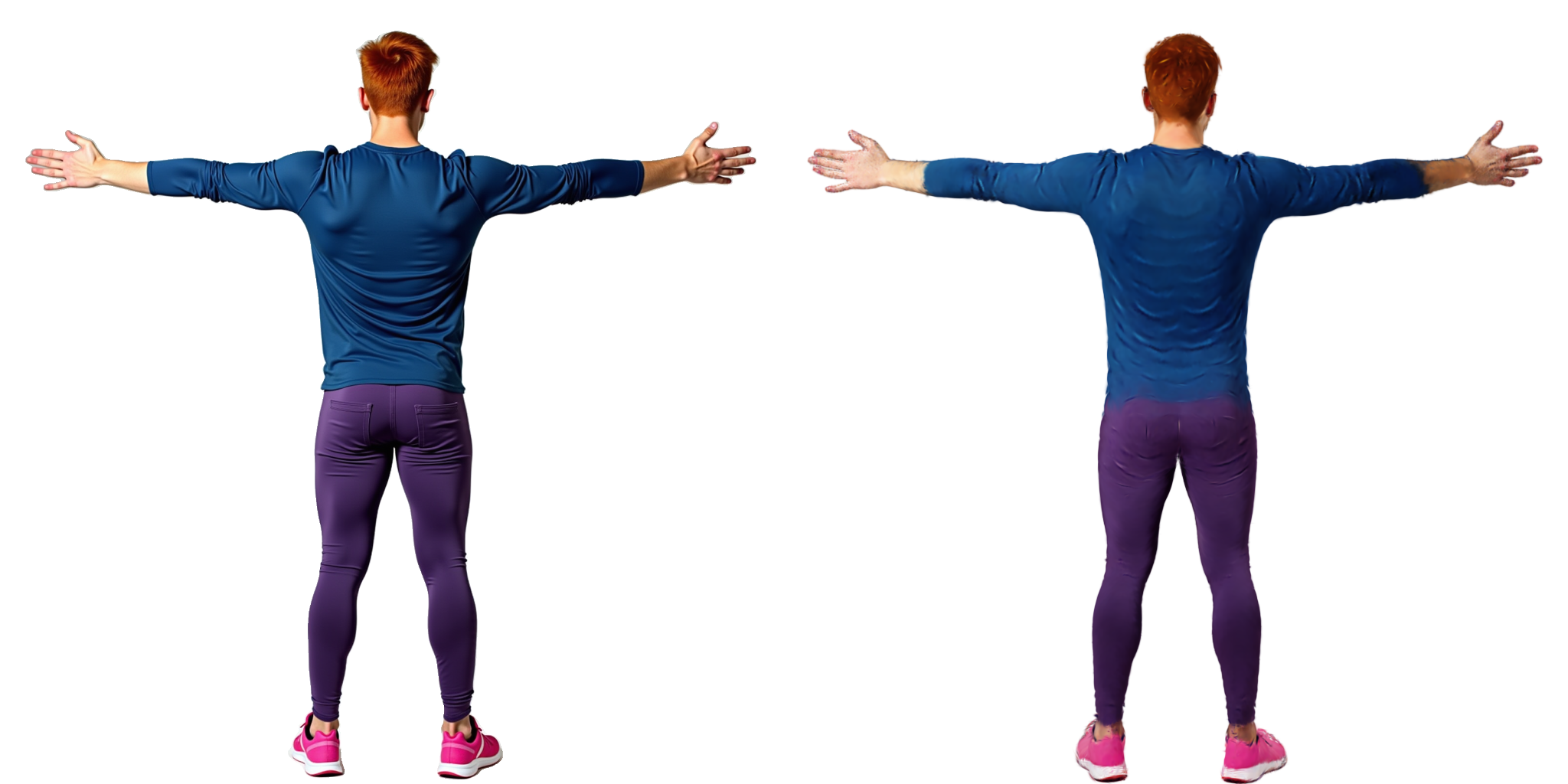}
    \includegraphics[width=0.16\linewidth]{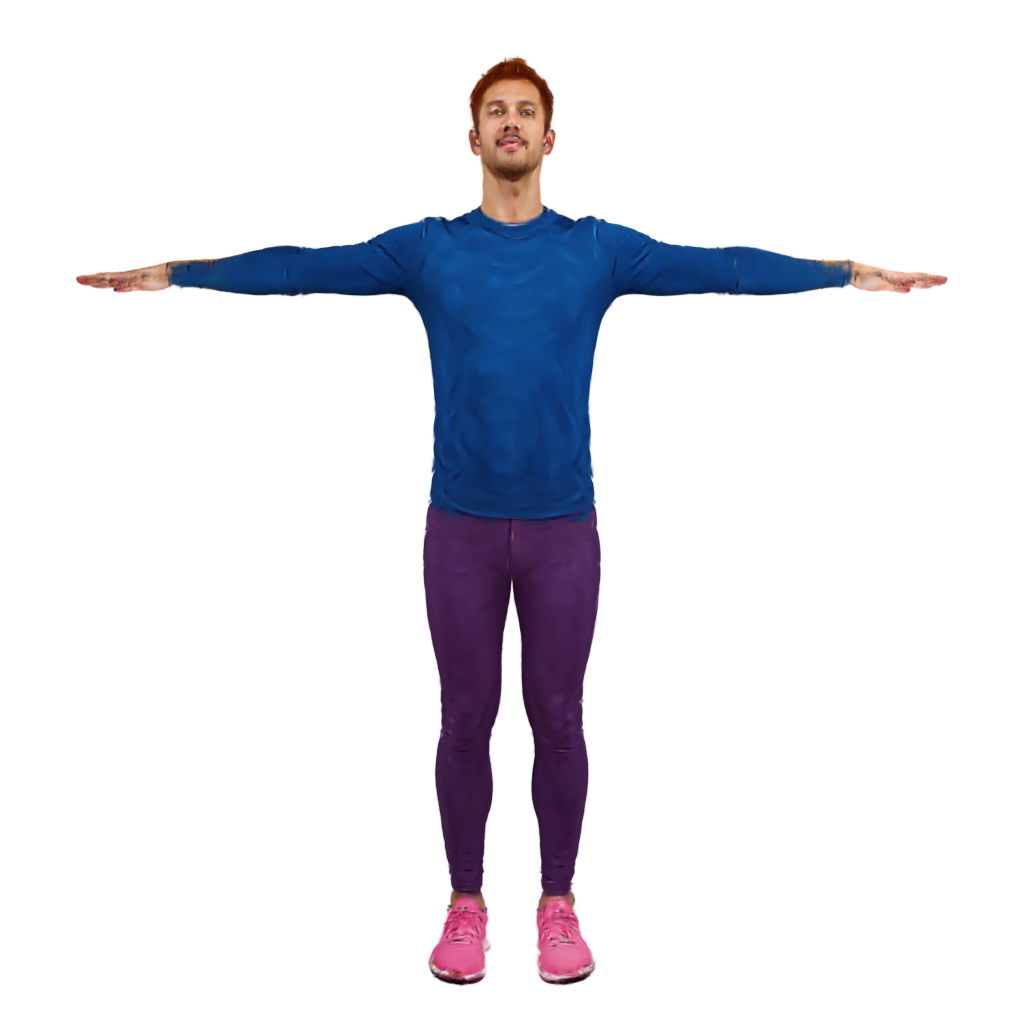}
    \includegraphics[width=0.16\linewidth]{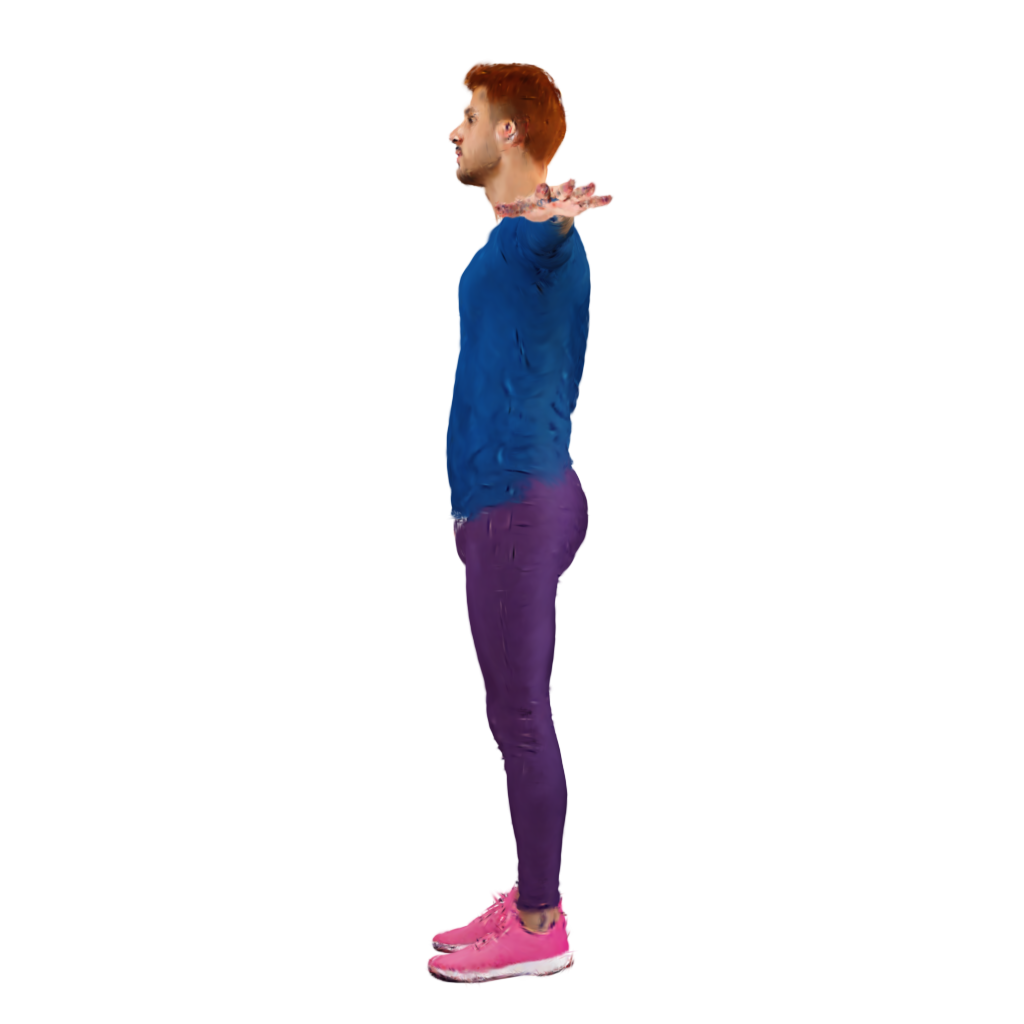}
    \includegraphics[width=0.16\linewidth]{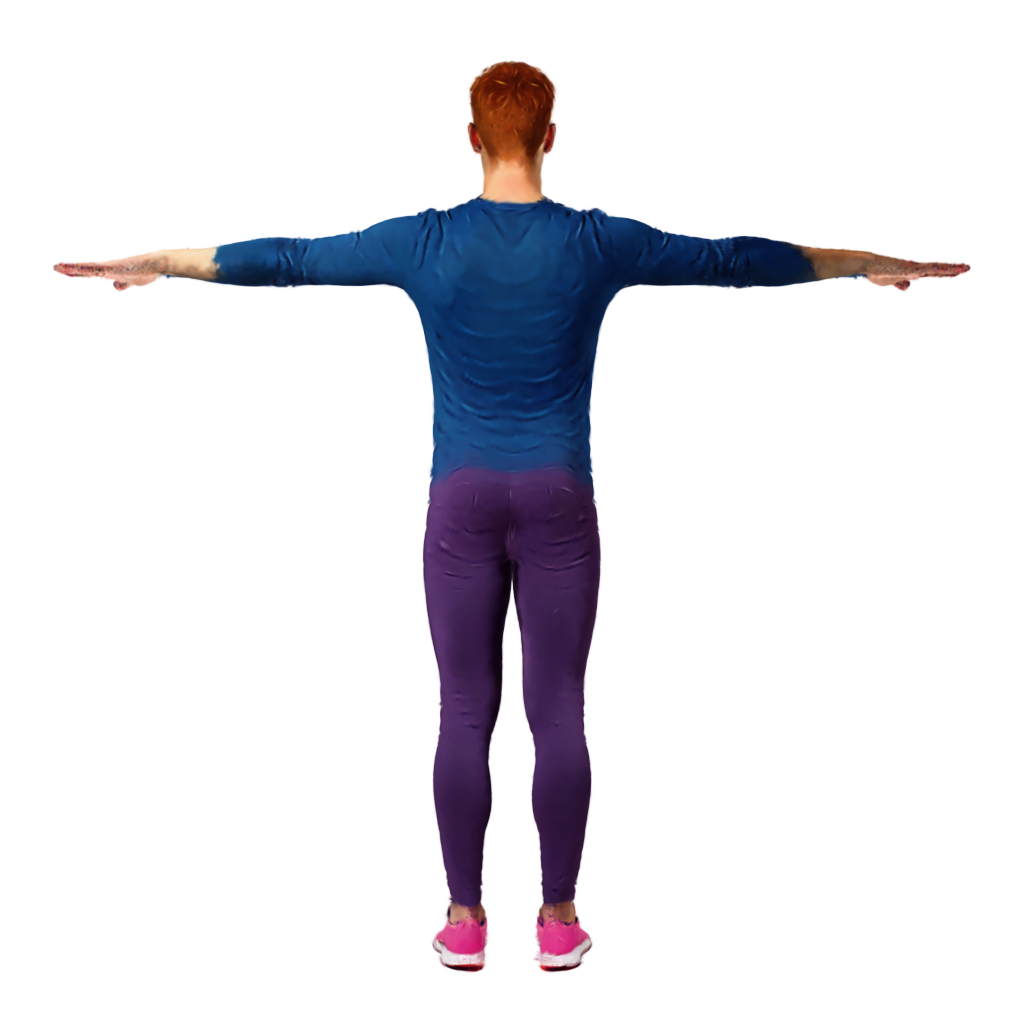}
    \includegraphics[width=0.16\linewidth]{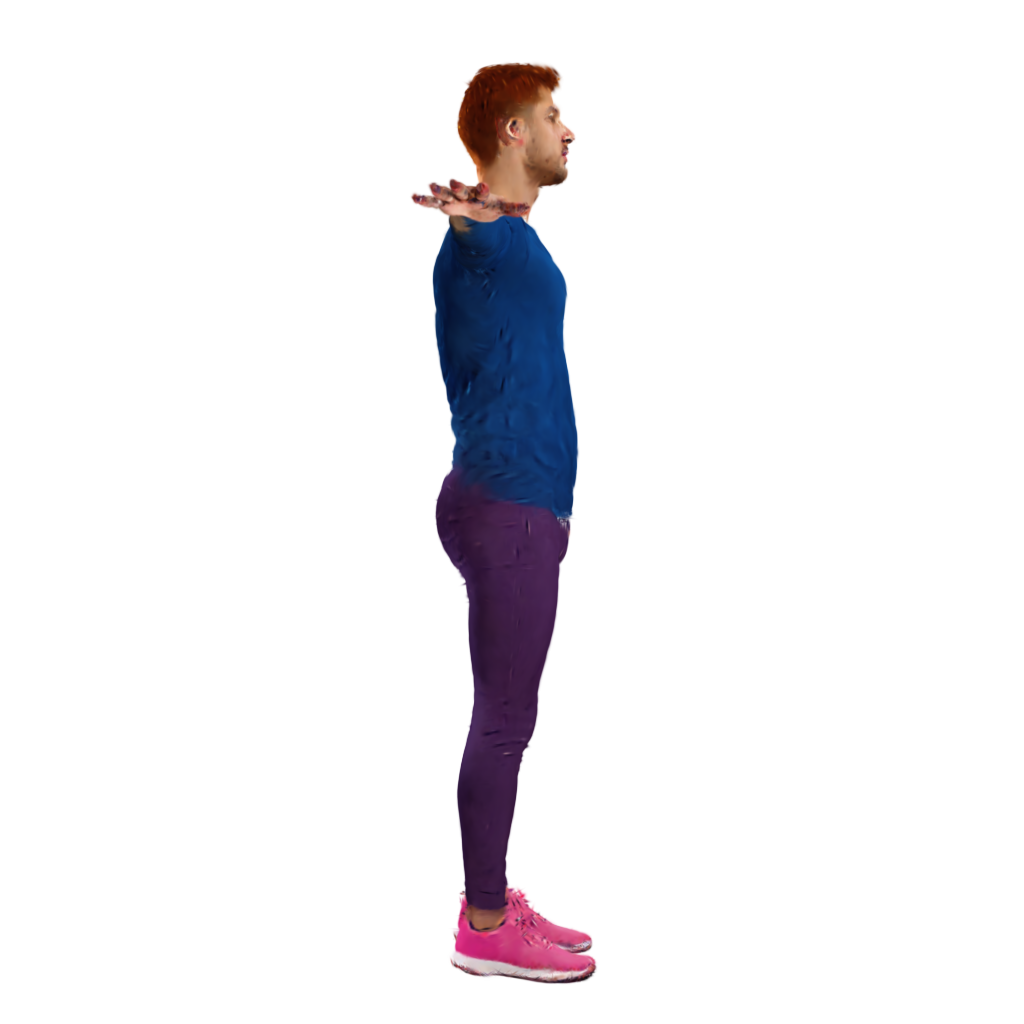}

    \includegraphics[width=0.31\linewidth]{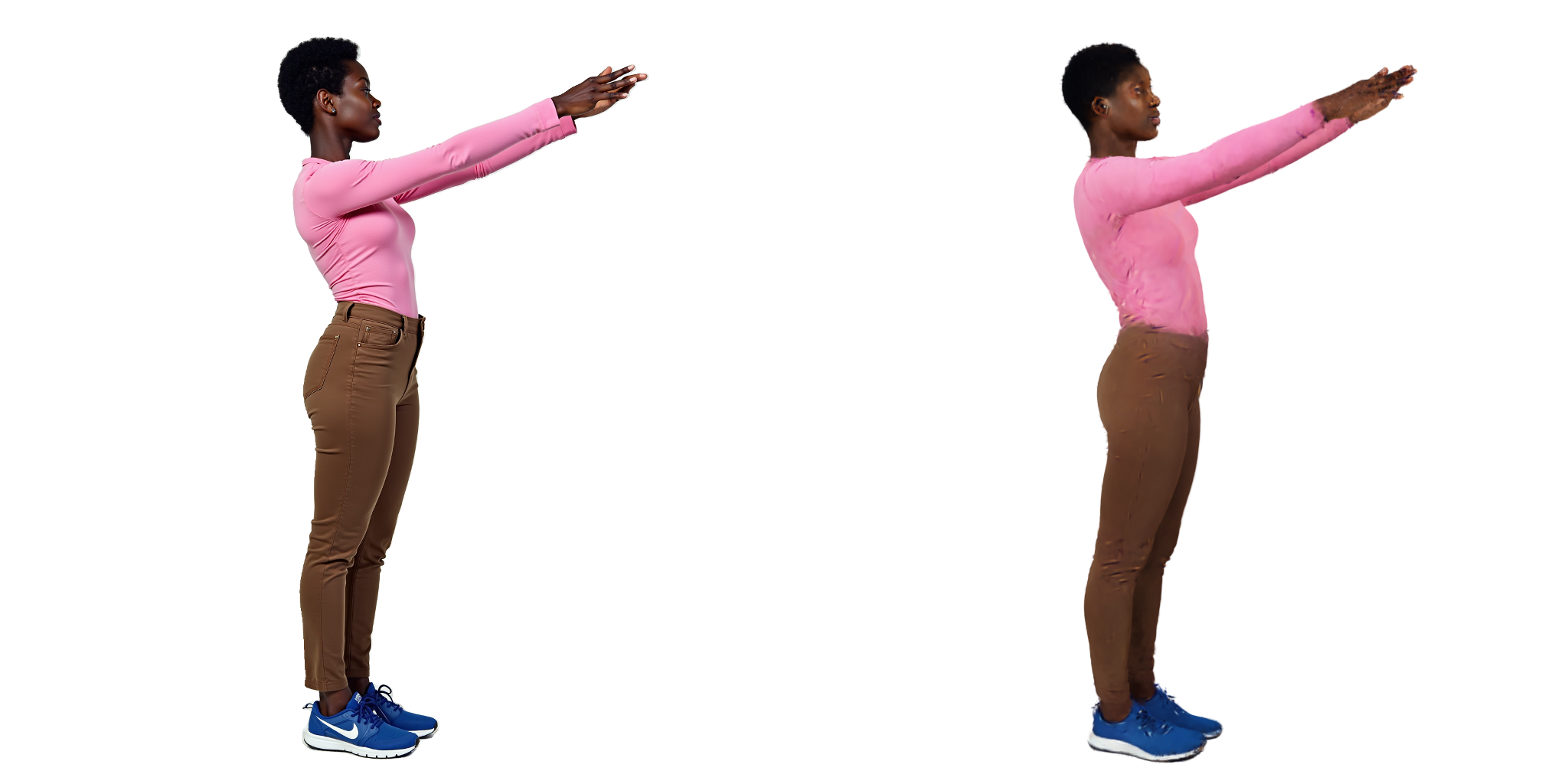}
    \includegraphics[width=0.16\linewidth]{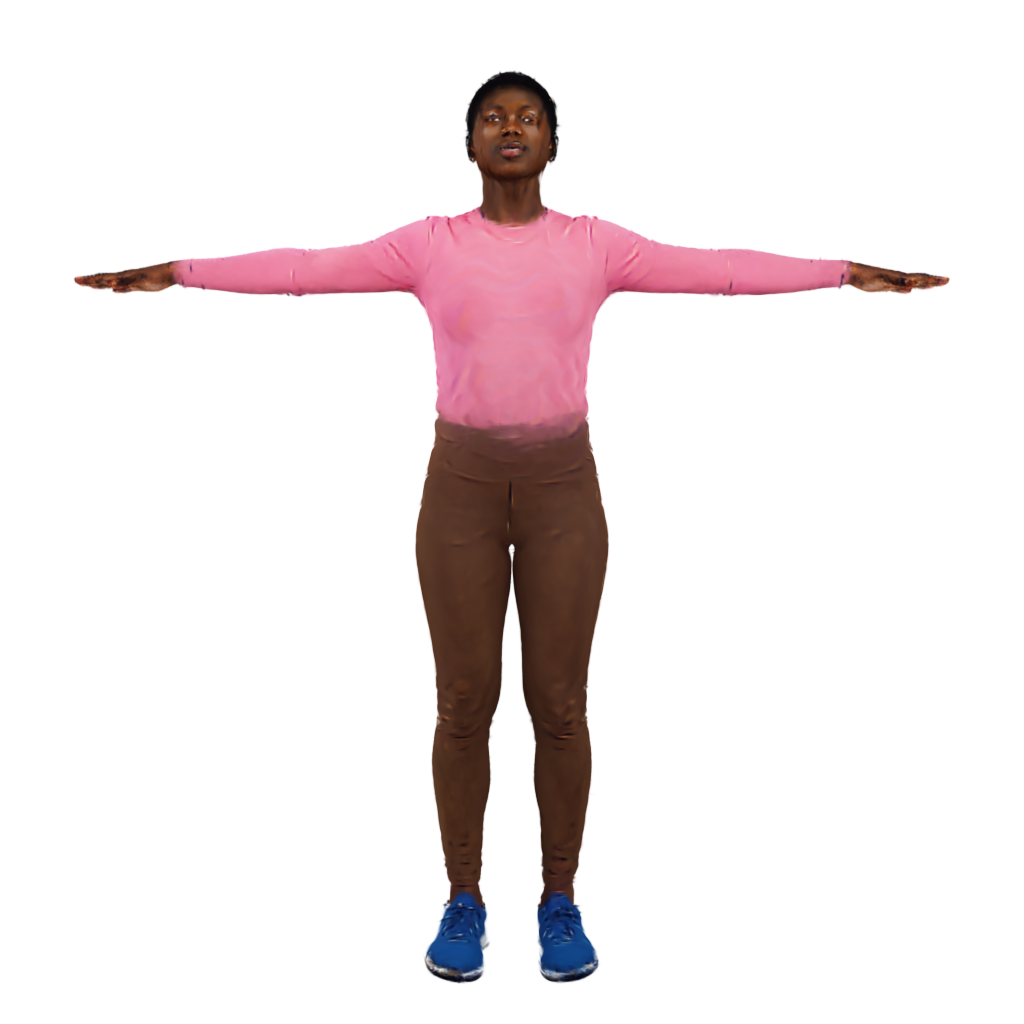}
    \includegraphics[width=0.16\linewidth]{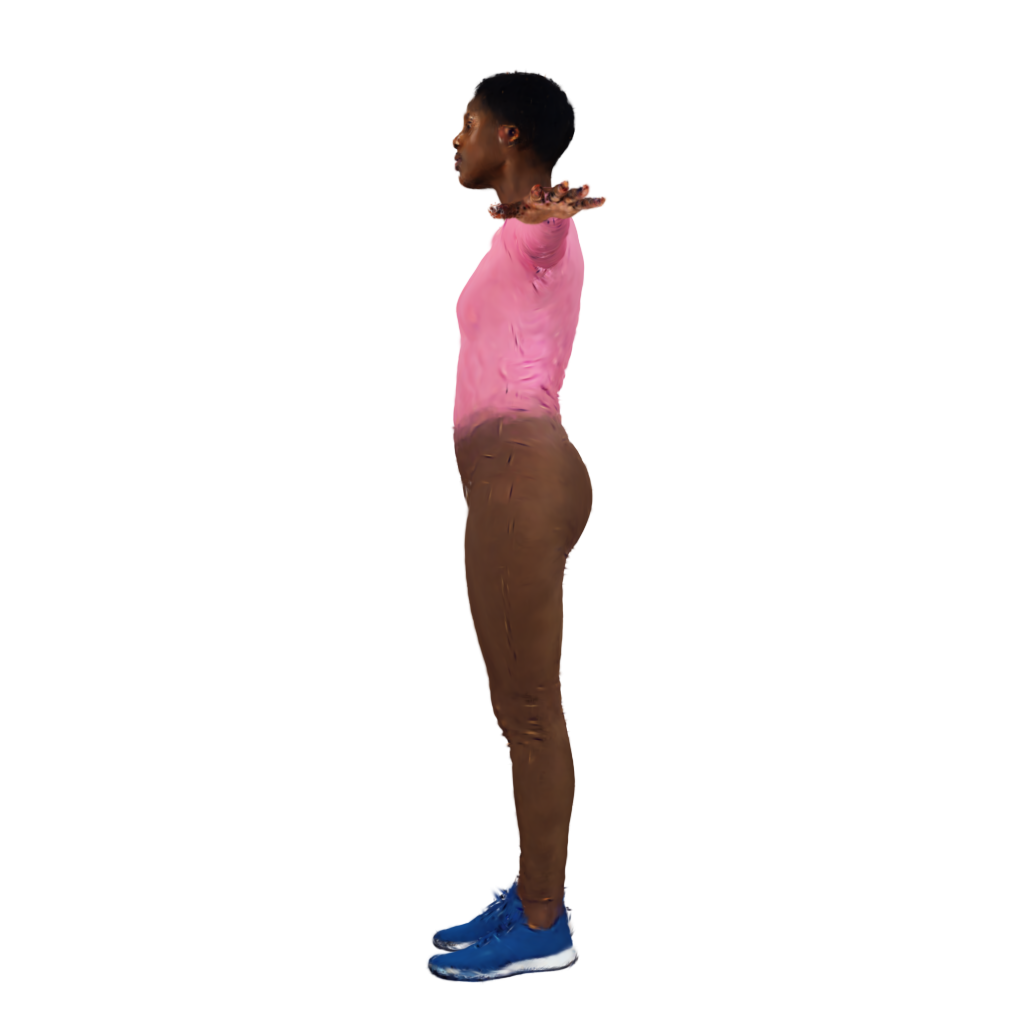}
    \includegraphics[width=0.16\linewidth]{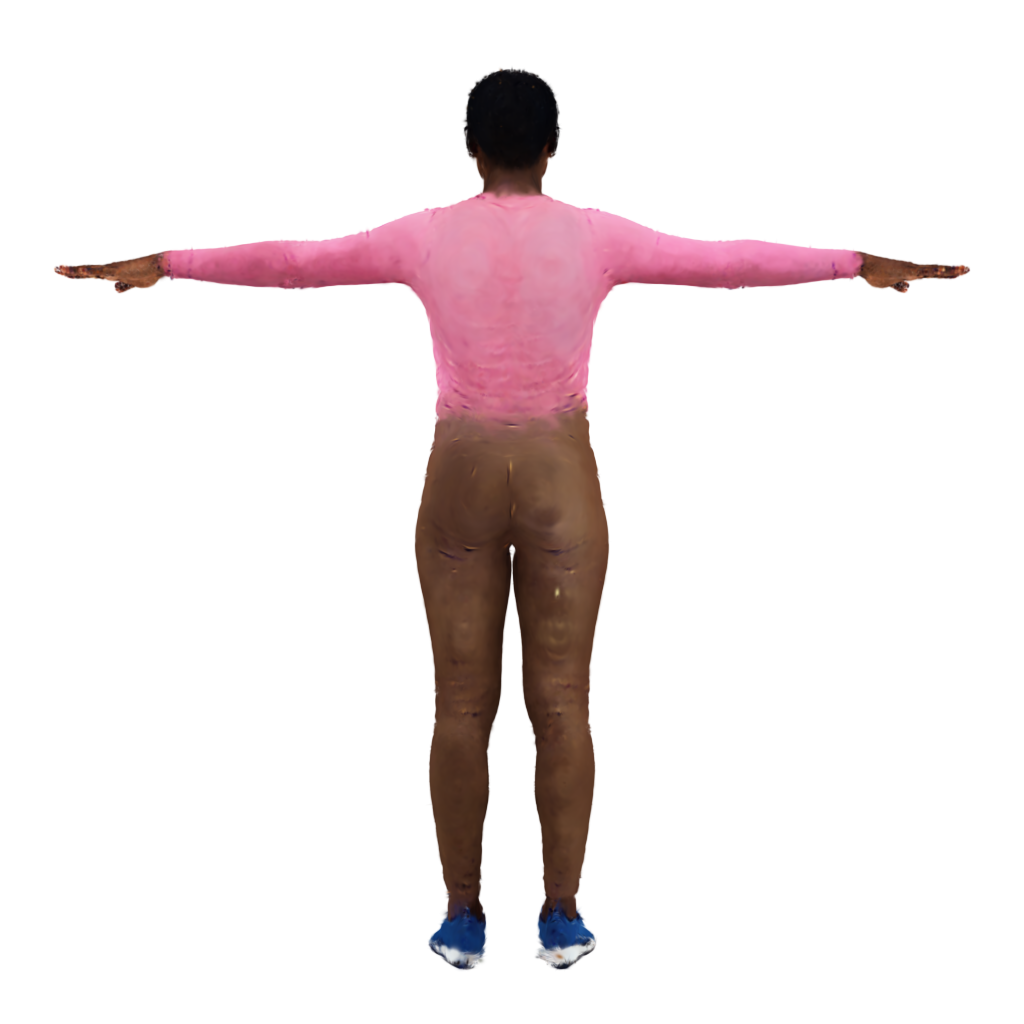}
    \includegraphics[width=0.16\linewidth]{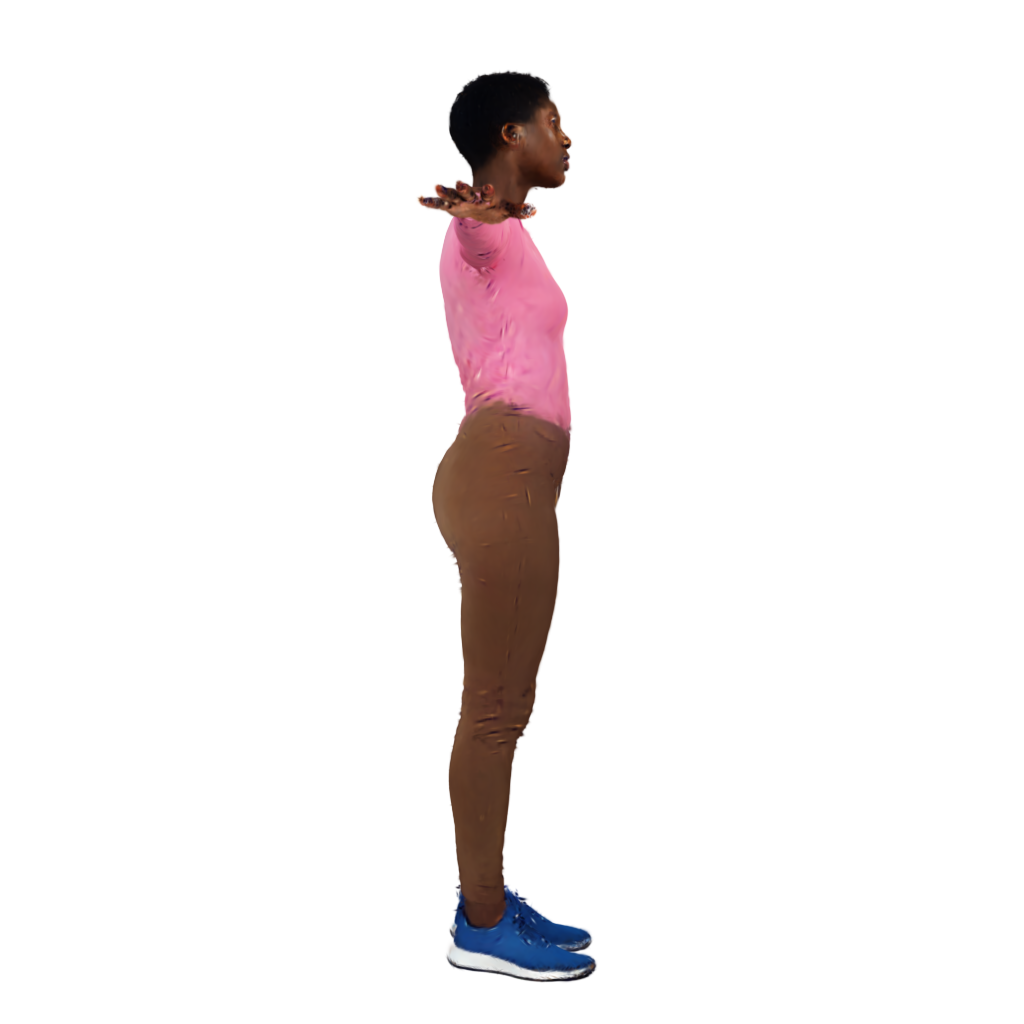}
        
    \caption{Novel view reconstruction. The first two images show the input image (left) and its 3D reconstruction in the same articulation pose (right). The following four images present novel view renderings of the reconstructed human in a canonical pose.}
    \label{fig:r1}
\end{figure*}
\begin{figure*}[t]
    \centering

    \includegraphics[width=0.35\linewidth]{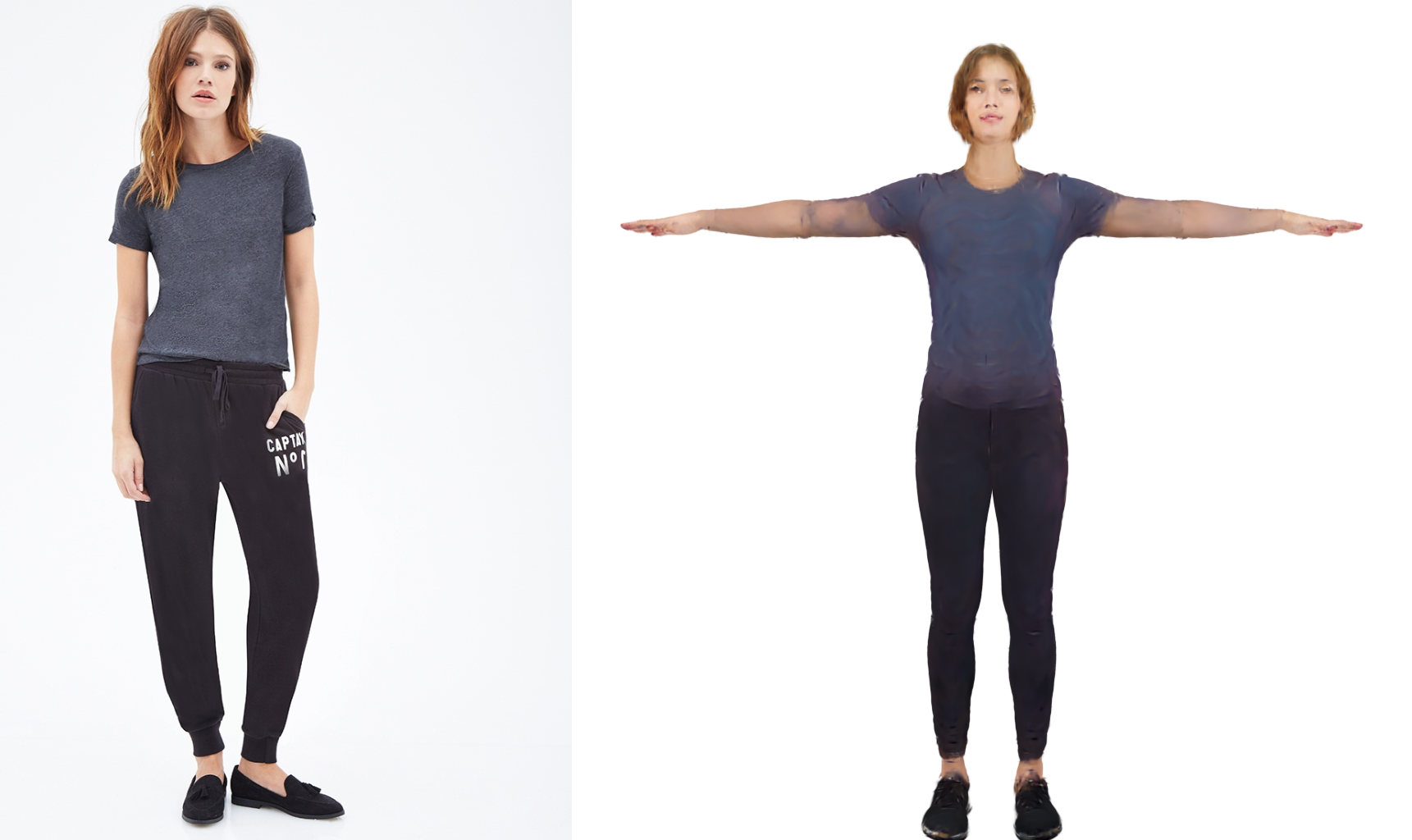}
    \includegraphics[width=0.21\linewidth]{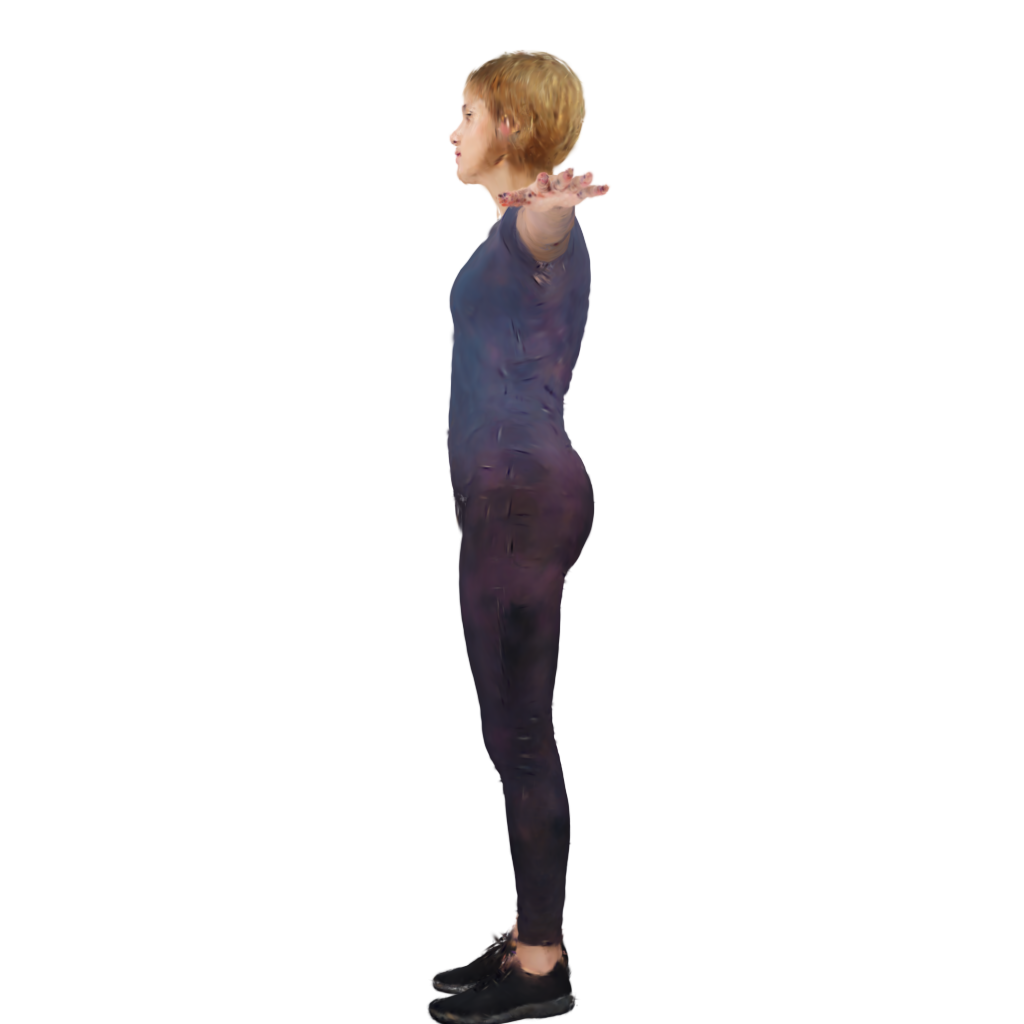}
    \includegraphics[width=0.21\linewidth]{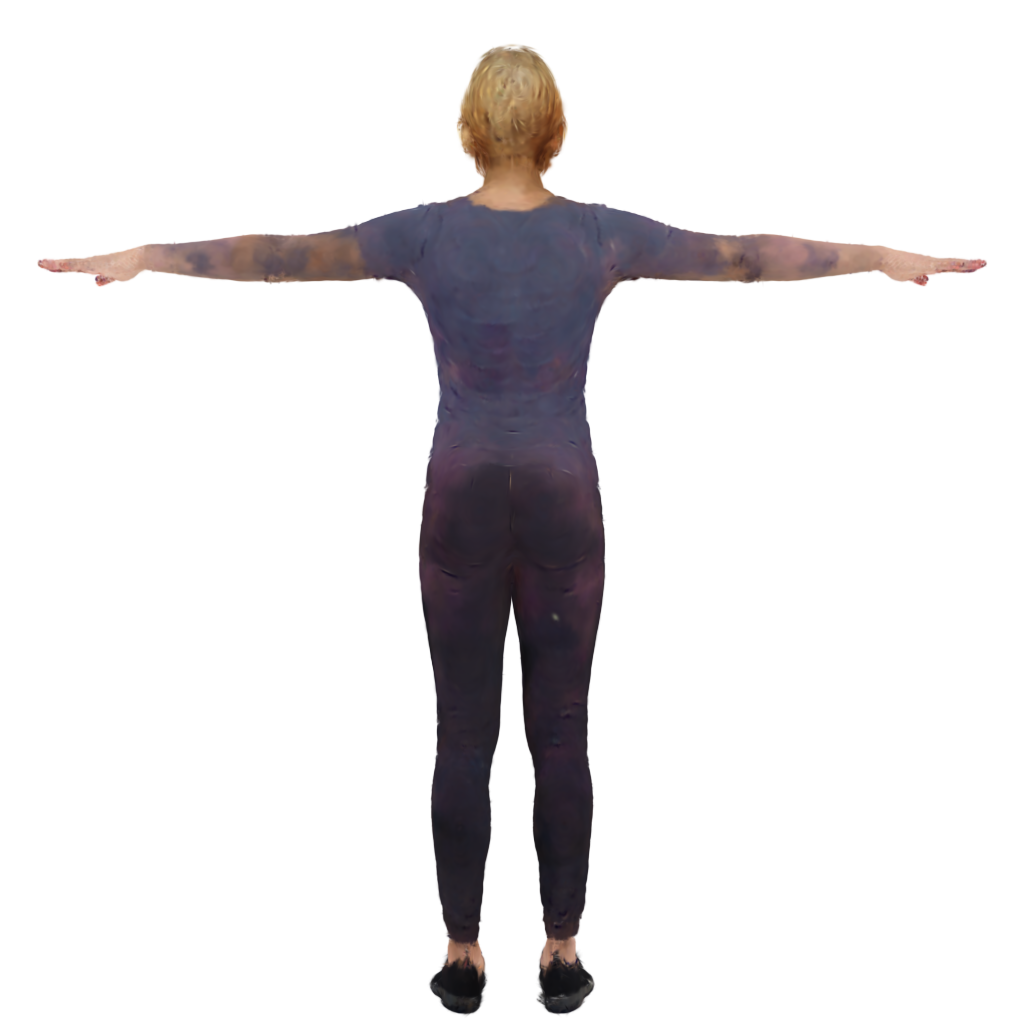}
    \includegraphics[width=0.21\linewidth]{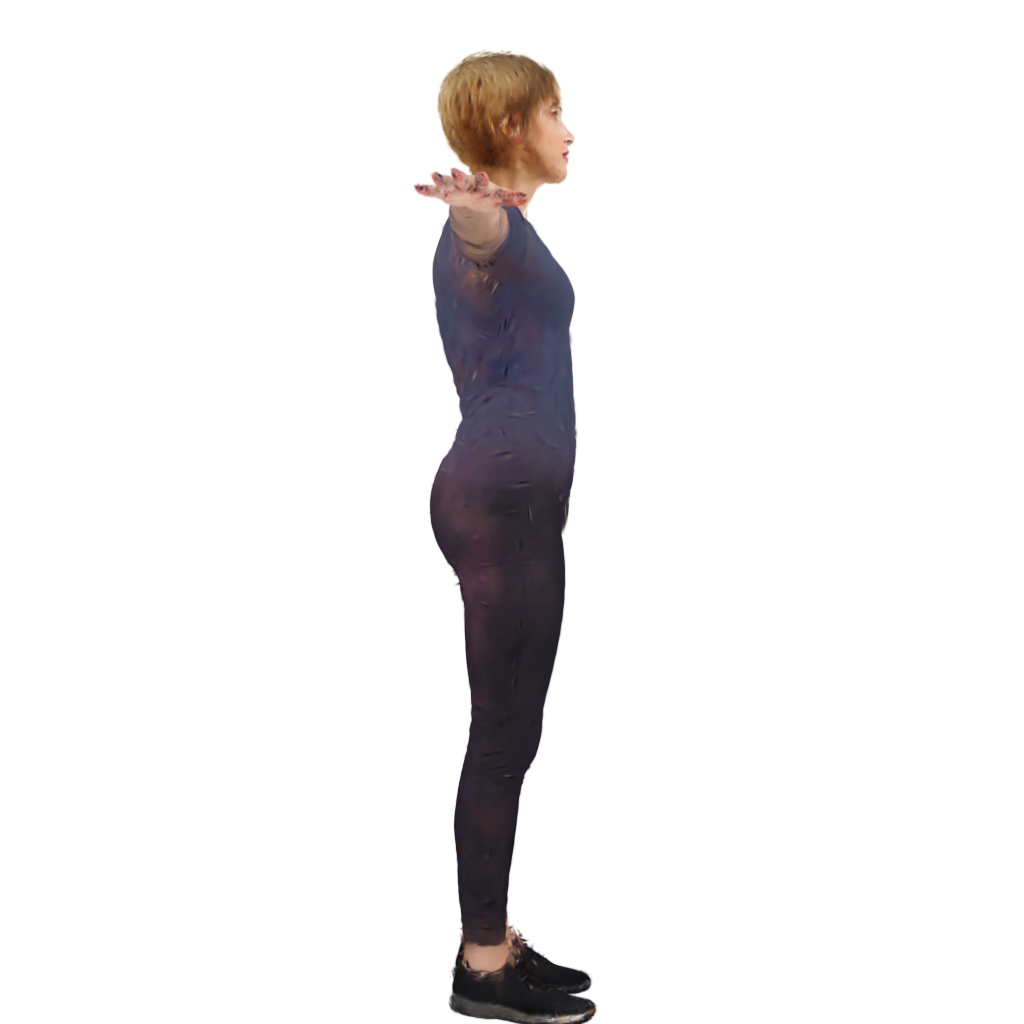}
    \caption{Novel view reconstruction on out-of-distribution images from the SHHQ~\cite{stylegan} dataset. The input image is on the left.}
    \label{fig:in-the-wild}
\end{figure*}

With keys obtained from the previous step, the self-attention is as follows:
\begin{equation}
    \mathbf{F}_X \leftarrow \text{softmax}(l_\text{PE}(\mathbf{F}_X)\mathbf{K}^\top) \: \mathbf{F}_X[K^\prime].
\end{equation}

\begin{equation}
  \centering
  \begin{aligned}
    (\mathbb{R}^{N\times h \times 1 \times \frac{d}{h}} \, \mathbb{R}^{N\times h \times \frac{d}{h} \times k})\, \mathbb{R}^{N\times h \times k \times \frac{f}{h}} \rightarrow \\ 
     \rightarrow \mathbb{R}^{N\times h \times 1 \times k} \, \mathbb{R}^{N\times h \times k \times \frac{f}{h}} \rightarrow\\
    \rightarrow \mathbb{R}^{N\times h \times 1 \times \frac{f}{h}} \rightarrow \mathbb{R}^{N\times f}
    \end{aligned}
\end{equation}



\section{Text Prompts Description}
The prompt given to the FLUX diffusion model was the following:
\begin{center}
    ``\emph{a real \{race\} \{gender\} with \{hair\} hair standing upright wearing tight \{color\} \{top\} and tight \{color\} trousers and \{color\} trainers, \{view\} view, arms stretched horizontally in a T-pose}''
\end{center}
The options for each variable in \{\dots\} are as follows:
\begin{enumerate}[noitemsep]
    \item {\bf race}: white, black, asian.
    \item {\bf gender}: man, woman.
    \item {\bf hair}: blonde, black, brown, ginger.
    \item {\bf tops}: long-sleeve t-shirt, t-shirt, long-sleeve shirt, shirt.
    \item {\bf view}: front, back, side.
    \item {\bf color}: red, brown, black, pink, yellow, blue, purple.
\end{enumerate}
Each attribute had a uniform probability of sampling. The examples of the dataset are in Fig.~\ref{fig:dataset}.

\section{Qualitative Results}
\begin{enumerate}[noitemsep]
    \item Humans generated by the diffusion model using the same prompt but different seeds are shown in Fig.~\ref{fig:same_prompt}.
    \item 3D human reconstructions from single images are shown in Fig.~\ref{fig:r1}.
    \item Reconstruction results on out-of-distribution data are shown in Fig.~\ref{fig:in-the-wild}.
    \item Novel views of the 3D humans generated via the diffusion model are shown in Fig.~\ref{fig:diff_novel_views}.
\end{enumerate}

\section{LLM Evaluation}
The large language models (LLM) we used to evaluate text-prompt alignment and image aesthetics are Gemini 2.5 Pro Preview~\cite{geminiteam2024geminifamilyhighlycapable}, Claude 3.7 Sonnet, Microsoft Copilot, and Grok 3.
The prompt we gave to the LLM for the evaluation is the following:
\begin{center}
    ``\emph{Evaluate the input images in terms of prompt alignment and aesthetic quality.
    For prompt alignment, use your own judgment to determine how well the image matches the given text prompt.
    The aesthetics is about overall image quality and how closely it resembles the appearance of a human, ignoring a human pose and background, including hands and face into consideration.
    Provide two scores between 0 and 1 for each image, 0 being the worst and 1 being the best.}''
\end{center}

\section{Ethical Statement}
The generated and rendered images, as well as manually defined text-prompt combinations, do not intend to discriminate or exclude other races, nationalities, or any other identities.
The idea of this work is to provide a theoretical background for generating diverse avatars and supporting anonymity by replacing a real appearance with a generated one. 
We acknowledge the potential for misuse of generative technologies and emphasize that this work is designed to promote positive applications, such as human representation in virtual reality.
Any similarities between the generated avatars and real-world individuals are purely coincidental and unintentional.
The generated process is entirely random and excludes any biases.
